\PassOptionsToPackage{table,xcdraw}{xcolor}
\documentclass[acmsmall,screen,nonacm]{acmart}
\AtBeginDocument{%
  }

\setcopyright{acmlicensed}
\copyrightyear{2025}
\acmYear{2025}
\acmDOI{XXXXXXX.XXXXXXX}

\acmJournal{THRI}
\acmVolume{37}
\acmNumber{4}
\acmArticle{111}
\acmMonth{2}

\usepackage{algorithm}
\usepackage{algorithmic}
\usepackage{amsmath}
\DeclareMathOperator{\sgn}{sgn}
\usepackage{multirow}
\usepackage{subfigure}
\usepackage{layouts}
\usepackage{siunitx}

\begin{document}

\title{Leveraging Passive Compliance of Soft Robotics for Physical Human-Robot Collaborative Manipulation}

\author{Dallin L. Cordon}
\authornote{Both authors contributed equally to this research.}
\email{cordond@byu.edu}
\orcid{0009-0004-9680-1482}
\author{Shaden Moss}
\authornotemark[1]
\email{engineering.moss@gmail.com}
\orcid{0009-0005-7874-5859}
\affiliation{%
  \institution{Brigham Young University}
  \city{Provo}
  \state{Utah}
  \country{USA}
}

\author{Marc Killpack}
\affiliation{%
  \institution{Brigham Young University}
  \city{Provo}
  \state{Utah}
  \country{USA}}
\email{marc_killpack@byu.edu}
\orcid{0000-0001-9372-104X}

\author{John L. Salmon}
\affiliation{%
  \institution{Brigham Young University}
  \city{Provo}
  \state{Utah}
  \country{USA}
}
\email{johnsalmon@byu.edu}
\orcid{0000-0002-8073-3655}

\renewcommand{\shortauthors}{Cordon and Moss et al.}

\begin{abstract}
  This work represents an initial benchmark of a large-scale soft robot performing physical, collaborative manipulation of a long, extended object with a human partner. The robot consists of a pneumatically-actuated, three-link continuum soft manipulator mounted to an omni-directional mobile base. The system level configuration of the robot and design of the collaborative manipulation (co-manipulation) study are presented. The initial results, both quantitative and qualitative, are directly compared to previous similar human-human co-manipulation studies. These initial results show promise in the ability for large-scale soft robots to perform comparably to human partners acting as non-visual followers in a co-manipulation task. Furthermore, these results challenge traditional soft robot strength limitations and indicate potential for applications requiring strength and adaptability.
\end{abstract}

\ccsdesc[500]{Computer systems organization~Robotic control}
\ccsdesc[500]{Computer systems organization~Robotic components}
\ccsdesc[500]{Computer systems organization~Robotic autonomy}
\ccsdesc[100]{Hardware~Physical verification}
\ccsdesc[500]{Hardware~Emerging technologies}
\ccsdesc[500]{Human-centered computing~Human computer interaction (HCI)}
\ccsdesc[500]{Human-centered computing~Collaborative and social computing}

\keywords{Physical human-robot interaction (pHRI), compliance, soft robotics, co-manipulation, pneumatics.}


\begin{teaserfigure}
    \centering
    \includegraphics[width=0.9\textwidth]{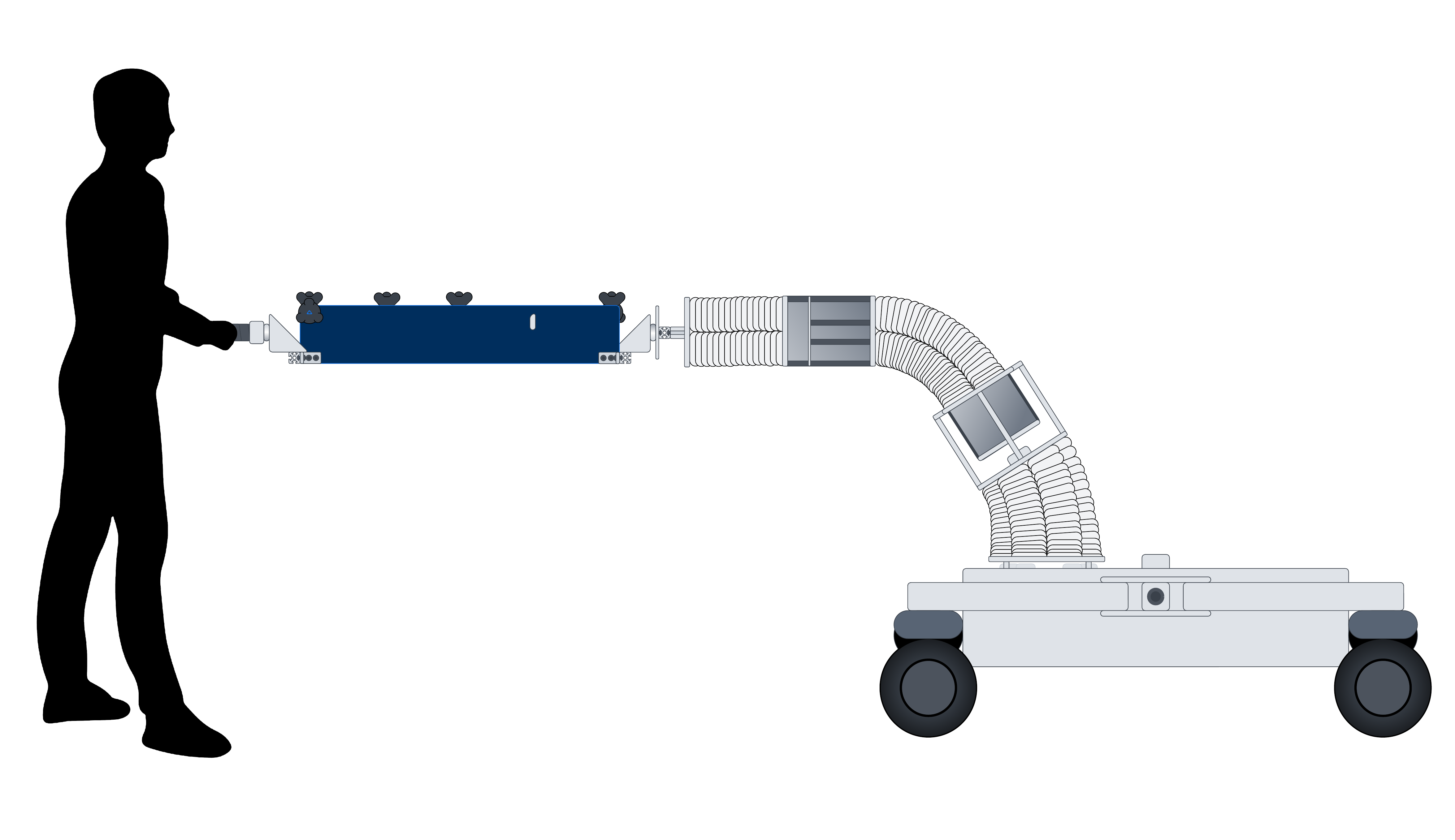}
    \caption{Collaborative manipulation between a human and a large-scale soft robot.}
    \Description[A human (left) collaboratively holding a stretcher-like object with a robot (right).]{Side view diagram showing a human (left) collaboratively holding a stretcher-like object with a robot (right). The robot consists of a soft arm with three continuum bellows joints mounted on a mobile base.}
  \end{teaserfigure}
\maketitle






\section{Introduction}

Collaborative manipulation of large, heavy, or unwieldy objects (defined in this paper as co-manipulation) has likely been a fundamental aspect of human interaction for millennia. As we seek to develop robotic helpers that operate more directly with humans, human-robot co-manipulation presents a unique challenge. For robots to work effectively with humans in shared object-carrying tasks, they must be designed to both complement human actions and align with human cognitive models so the robots' interactions become intuitive and natural for their human partners. Furthermore, it is imperative that robots ensure the safety of their human counterparts by adjusting to unpredictable movements and disturbances while avoiding excessive force.

The field of soft robotics has been presented as a highly viable solution to enable effective and safe physical human-robot interactions. The inherent compliance of soft robots is ideal for safe interactions as their flexible and adaptive structures can absorb impact and reduce the risk of injury during contact. This makes them particularly well-suited for tasks that require close physical proximity to humans, where safety and adaptability are paramount. 
However, much of the work in soft robotics has been concentrated on silicone-based robots, which tend to have low force output and limited power. Other work has focused on grippers designed to manipulate fragile objects of low mass and volume \cite{shintake_soft_2018}, truss-based robots \cite{usevitch_untethered_2020}, expanding robots engineered to navigate cluttered environments \cite{hawkes_soft_2017}, and wearable robots \cite{zhu_soft_2022}. Few researchers have demonstrated soft robots with capabilities for external, heavy-duty collaborative manipulation tasks with humans. 


\begin{figure}[!t]
    \centering
    \includegraphics[width=0.7\linewidth]{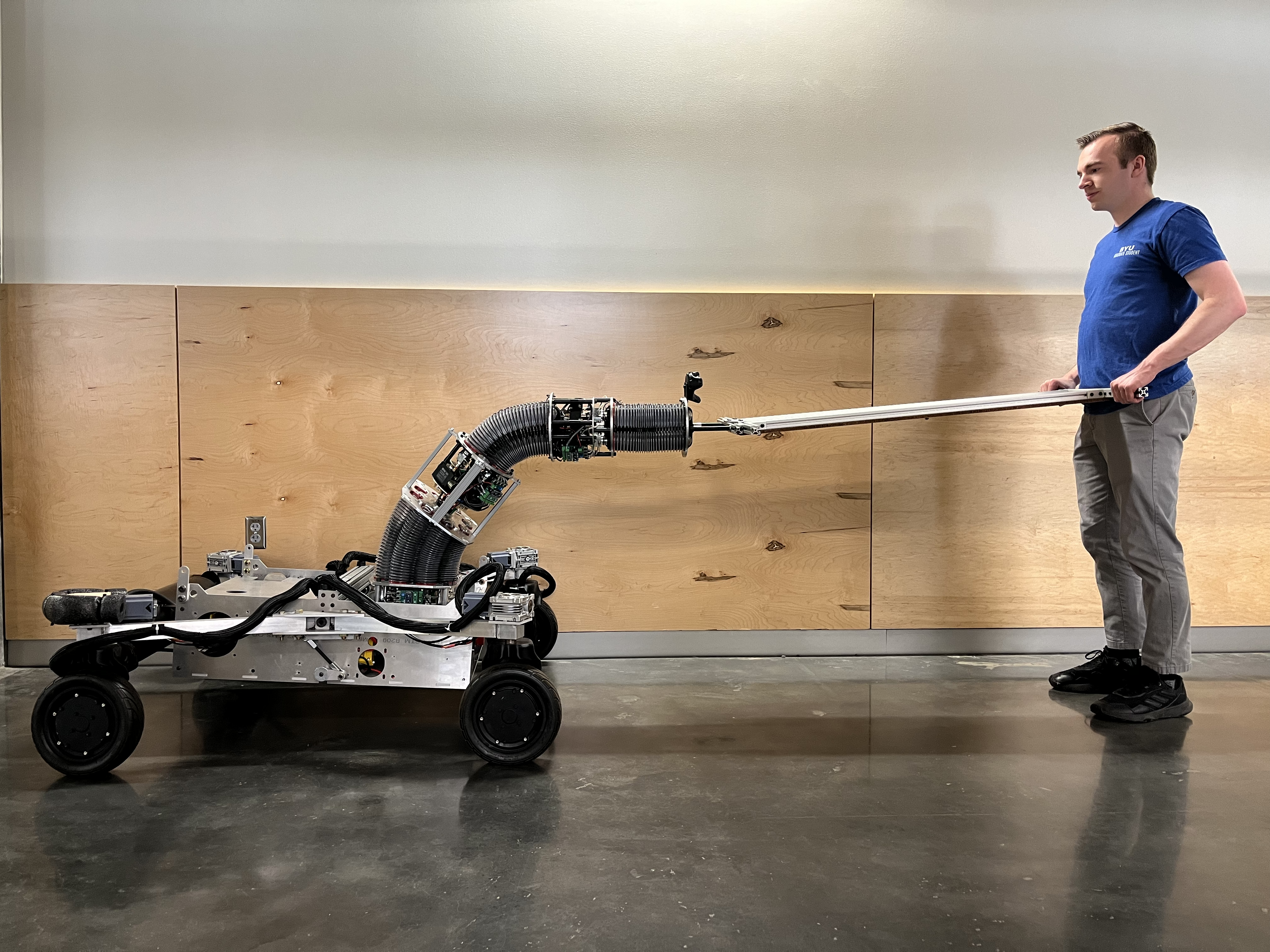}
    \caption{Physical human-soft-robot co-manipulation with a passively compliant robot arm.}
    \label{fig:comanip}
    \Description[Photograph of a human (right) collaboratively holding a stretcher-like object with a robot (left).]{Photograph of a human (right) collaboratively holding a stretcher-like object with a robot (left) in a hallway. The robot consists of a soft arm with three continuum bellows joints mounted on a mobile base.}
\end{figure}

This work demonstrates the potential of large-scale soft robotics for effective human-soft-robot interaction through a novel system: a passively compliant soft-robotic manipulator mounted on a custom omni-directional mobile base (Figure~\ref{fig:comanip}). We evaluate the system's performance through human-soft-robot co-manipulation tasks involving a long, extended object and compare that performance to previous human-human co-manipulation experiments with similar payloads \cite{shaw_thesis_2024, freeman_classification_2024}. 

In our study, nineteen individuals were recruited to perform planar translations (forward and back, side-to-side, and diagonal) of the co-manipulation object (CMO) while immersed in a virtual environment. The translational tasks were chosen to match a subset of the tasks performed by Shaw \cite{shaw_thesis_2024} to enable direct performance comparison. Participants completed sixteen sets of randomized translational tasks, each under different robotic parameters. The NASA Task Load Index (NASA-TLX) was administered after each set to assess perceived workload. While the full parameter study is being analyzed separately, this work focuses on four sets using consistent robot parameters from which we perform comparisons with previous human-human studies. 

Interaction between the robot and the human took place solely through the CMO, with no direct physical contact. The human operator possessed exclusive knowledge of the object's desired trajectory and environmental constraints allowed for the existence of a direct, feasible path from the initial position to the desired goal. The robotic arm actively contributed to support some of the weight of the object throughout the co-manipulation task. 

Survey questions from Freeman et al. \cite{freeman_classification_2024} were administered to participants both before and after the experiment. Shaw \cite{shaw_thesis_2024} employed a subset of Freeman's post-study questions in their subsequent research, enabling qualitative analysis across all three studies.

In short, the main contributions of this work are:

\begin{itemize}
    \item The development of a novel mobile soft-manipulation robotic platform for human-soft-robot collaborative manipulation.
    \item A benchmark study for physical human-soft-robot co-manipulation with a soft robot arm.
    \item An analysis of our robot partner's co-manipulation performance relative to human partner performance for similar tasks.
\end{itemize}

We seek to lay the foundation for future work focused on optimizing the control and coordination of these soft mobile manipulators for co-manipulation tasks in order to further improve user acceptance and team efficacy.

The remainder of this paper is structured as follows: Section \ref{sec:background} provides background information and context for our work. Section \ref{sec:method} details our methodology, including the system's hardware design - specifically the mobile base and robotic arm - along with the control architecture and the design of our experiment. Finally, Sections \ref{sec:results} and \ref{sec:disc} present the experimental results and their associated discussion.


\section{Background}
\label{sec:background}

In co-manipulation tasks with rigid manipulators, compliance-based control is essential to address the complex interplay between the manipulator and the object. Challenges may arise from uncertainties in both the robot's kinematic model and the grasping technique. These uncertainties can generate unexpected internal forces between the robots' end-effector and the manipulated object, potentially compromising task performance and safety. Impedance control---and its counterpart, admittance control---as pioneered by Hogan \cite{hogan_impedance_1984}, has emerged as a prevalent approach in co-manipulation control strategies, enabling robots to effectively respond to positional deviations or interaction forces at the end-effector with the appropriate mechanical compliance \cite{keemink_admittance_2018, song_tutorial_2019}.

Ikeura \cite{ikeura_variable_1995} presented one of the first works in co-manipulation using a variable impedance controller. Since then, researchers have expanded on this concept to enhance co-manipulation performance. Duchaine et al. introduced an approach implementing force derivatives for human intention estimation and showed that velocity-based variable impedance control outperformed position-based approaches for human-soft-robot collaboration \cite{duchaine_general_2007}. 
Gopinathan et al. explored adapting stiffness online using force measurements from initial participant tests, demonstrating that adaptive stiffness was preferred over static settings in more complex tasks \cite{gopinathan_user_2017}. Though some studies initially focused on single degree of freedom (DOF) tasks, impedance-based control has been successfully scaled up to 4 DOF \cite{lecours_variable_2012}, and even to 6 DOF tasks, where Whitsell and Artemiadis demonstrated a system that enables the manipulator to actively control the degrees of freedom of a small, co-located object that the human is not actively manipulating \cite{whitsell_physical_2017}. For an extensive review of variable impedance control applied to human-robot interactions, see \cite{sharifi_impedance_2022}.



While compliance-based control has been a central strategy in co-manipulation, researchers have also explored alternative approaches that bypass the complexities of impedance tuning altogether. Many of these approaches focus on purely robotic teams and omit the use of a full manipulator. Notably, caging involves mobile robots surrounding an object and utilizing formation-based techniques for transportation \cite{song_potential_2002, wang_object_2002, pereira_decentralized_2004, habibi_distributed_2015, fink_multi-robot_2008, wan_cooperative_2012, hichri_design_2019}. Other strategies involve estimating the pose \cite{pereira_object_2010} or relative motion \cite{rauniyar_mewbots_2021, elwin_human-multirobot_2023} of the co-manipulated object (or its point of contact) and adjusting motion accordingly. Methods like caging and pose estimation simplify object movement control by avoiding precise force feedback, but they limit manipulation to basic transport, lacking the nuanced control needed for human interactions. However, in a similar vein to our proposed research, an early approach introduced compliance by incorporating a proposed viscoelastic trunk, represented by a spring and damper, between the mobile base and the manipulator arm \cite{lim_collision-tolerant_1998}.


Soft robotics presents a compelling alternative by introducing passive compliance directly into the system, offering natural adaptability and enhanced safety during human-robot interaction. This compliance minimizes the need for complex control strategies to manage force interactions, as the flexible structure can absorb and distribute forces more effectively than rigid systems \cite{rus_design_2015}. However, completely soft robots are unlikely to fully realize the potential of co-manipulation, especially as soft mobile robots are still in their relative infancy and unsuited for large-scale mobile manipulation due to limited size, payload capacity, and maneuverability \cite{onal_soft_2017, drotman_electronics-free_2021, li_fast_2018, foster-hall_soft_2024}. Instead, a hybrid approach using a traditional robot base with a soft manipulator \cite{stokes_hybrid_2014, hyatt_configuration_2019, liu_underwater_2020} combines the stability and precision of a standard mobile platform with the compliance and safety of a soft robotic arm.


A number of studies focused on human-human co-manipulation, provide valuable benchmarks and comparison points for assessing the collaborative performance of our robotic platform. 
Freeman et al. had participants carry a 27 kg stretcher-like object through a series of obstacles under various task-modi \cite{freeman_classification_2024}. Shaw \cite{shaw_thesis_2024} conducted similar experiments with participants wearing VR headsets performing 1-3 DOF tasks in leader-follower, leader-leader, and leader-follower-follower dyad and triad configurations where the leader is aware of the target position and orientation of the object and the follower is not.
These studies collected data on task completion times and participant feedback, establishing a reference for evaluating the effectiveness of human-robot co-manipulation systems and acting as a critical precursor. By comparing our results to these human-human interactions, we aim to better understand the extent to which soft robotic systems can replicate the nuanced dynamics of human teamwork.



Although there has been substantial work dedicated to robotic co-manipulation with humans, large-scale, mobile, soft robotic manipulators have yet to be considered for such tasks. This work focuses on bridging the gap between physical human-robot interaction via co-manipulation tasks and the growing field of soft robotics.

\section{Methodology}
\label{sec:method}
Building upon the potential of hybrid soft robotic systems, our study aims to evaluate the performance of a novel co-manipulation platform. The following methodology details our robotic platform as well as the design of our experiment, which tests the compliance, adaptability, and collaborative capabilities of our proposed large-scale soft robotic platform.

Due to the novelty of soft robotic co-manipulation, our high-level approach adopts simpler, passive strategies. 
The initial focus is on planar, 2-DOF motions, setting aside the complexity of 6-DOF operations for future work.
The robot platform is rigidly connected to the co-manipulation object and is outfitted with force/torque sensors at the connection point to the end effector. 
Loading and unloading of the object is not considered. 
External line-of-sight sensors are used to estimate joint angles and provide the arm with proprioception and control. 
We anticipate enhancements to refine these necessary simplifications over time.

\subsection{Physical System Design}
\label{sec:robot-platform}

The robotic platform consists of a pressure-driven bellows-joint arm mounted onto an omni-directional mobile base. The end effector is rigidly fixed to the co-manipulation object.

\subsubsection{Mobile Base}
\label{sec:mobile_base}

\begin{figure}
    \centering
    \includegraphics[width=0.7\linewidth]{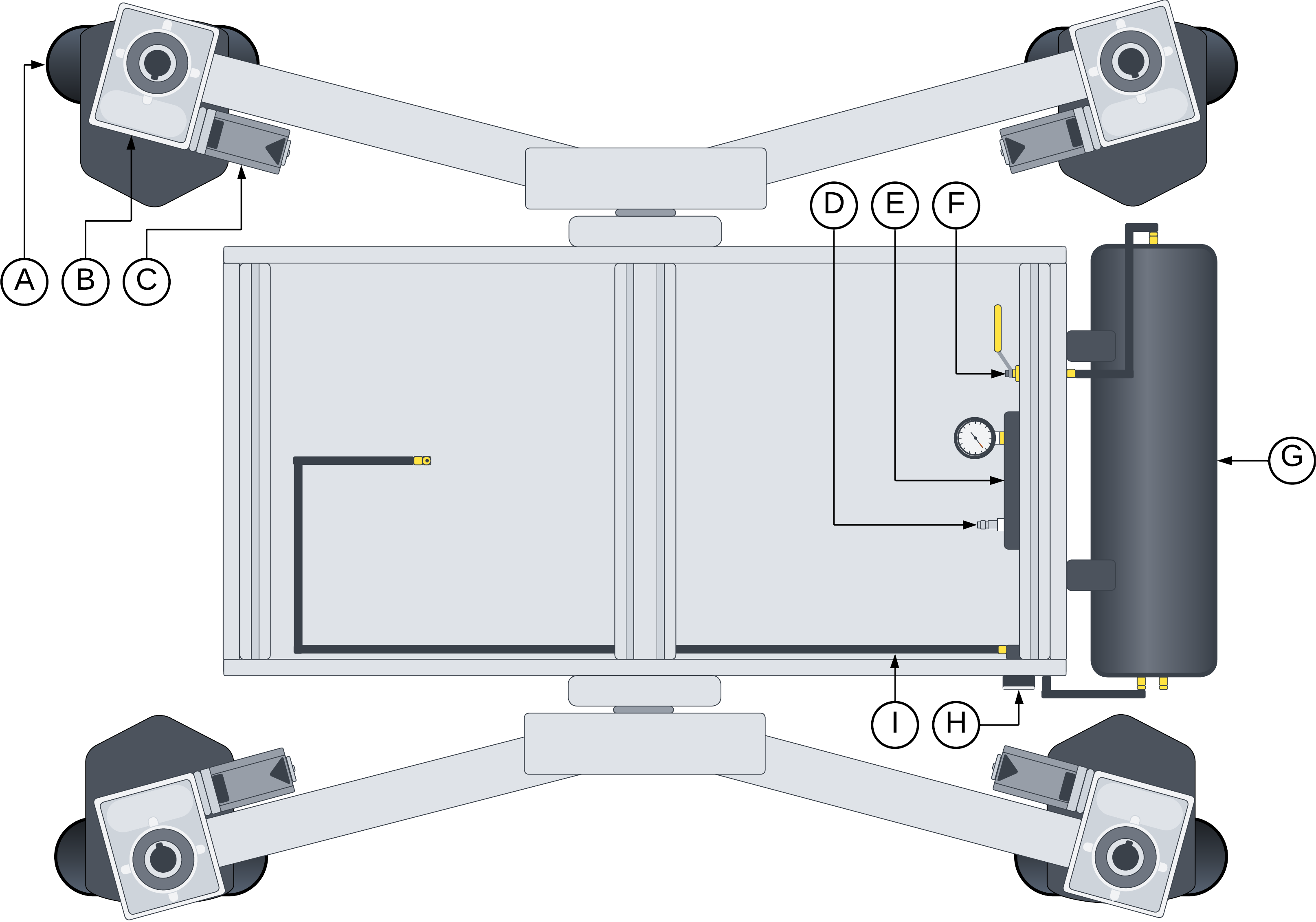}
    \caption{Top-down, annotated view of the mobile base platform and the pressure system. Annotated components include: (A) wheels and hub motors, (B) 90$^{\circ}$ castor gearbox, (C) castor motor, (D) pressure manifold intake, (E) primary pressure manifold assembly, (F) pressure tank access valve, (G) high pressure storage tank, (H) pressure regulator, and (I) pressurized air transmission line to arm.}
    \label{fig:mobile_base}
    \Description{Top-down diagram of a robot. The robot has four struts extending from the center of the rectangular frame, with the wheel assemblies--consisting of powered castors and wheels--mounted on the ends. Mounted to the back of the robot frame is a pressure tank along with associated valves, pressure lines, and regulators.}
\end{figure}

The mobile base, seen in Figure~\ref{fig:mobile_base}, is a wheeled omni-directional platform developed by a team of undergraduate and graduate students in the BYU Robotics and Dynamics (RAD) Lab. Built with an extruded aluminum frame reinforced by water-jet aluminum panels, the base achieves omni-directional movement through four hub motors and motorized castors. This configuration enables simultaneous forward/backward motion, side-to-side translation, and rotation.


The system uses ODrive V3 motor controllers for low-level PID control with encoder feedback. An Alienware M15 laptop with an NVIDIA GeForce RTX 3080 GPU runs a custom ROS package that converts velocity vector inputs into CAN Bus commands for the motors and castors. The laptop also handles data collection, including audio from a USB microphone and force-torque measurements from three ATI Mini45 sensors. While not used in this study, the base includes a ZED camera for SLAM, computer vision, and IMU-based tracking to support future optimal control experiments.


The mobile base's pneumatic system includes a manifold, pressure tank, and pressure regulator that supplies compressed air to the manipulator. An external air source connects to the manifold. When the tank access valve is opened, the compressed air fills the pressure vessel located at the base's rear. The tank stores compressed air from the external source--acting as a reservoir or fluid capacitor--before passing through a pressure regulator. The pressure regulator limits air pressure to around 200 \unit{\kPa} before routing the compressed air to the arm. At the base of the arm, the air is distributed to each of the individual valves that control the manipulator's bellows chambers.

The base has demonstrated capabilities of carrying 80 kg payloads at speeds up to 5 m/s, though these exceed the requirements of this work. For this study, the maximum velocity was limited to 0.5 m/s for safety.
    

\subsubsection{Arm}
\label{sec:arm}

The arm design, described in detail by Johnson et al. \cite{johnson_baloo_2025}, features three pneumatically actuated bellows joints connected by rigid links containing control valves. We model the kinematics of each joint as having two rotational degrees of freedom, actuated through four bellows chambers (two antagonistic pairs). While the base joint contains eight chambers, it is likewise managed by four valves--two chambers per valve--and a single valve for each quadrant. The joints use Enfield Technologies valves of decreasing size (LS-V25s, LS-V15s, and LS-V05s from base to end) and Pneubotics-manufactured blow-molded PETE plastic bellows. A central spine of braided polyurethane-coated aramid rope maintains joint compression.


The PneuDrive distributed pressure control system manages chamber pressures \cite{johnson_pneudrive_2024}, while joint angles are estimated using end-point orientations under a constant-curvature assumption \cite{hyatt_model_2020, jensen_tractable_2022, allen_closed_2020}, tracked via VIVE motion sensors.



\subsubsection{Object}
\label{sec:object}

\begin{figure}
    \hfill
    \subfigure[Top-down and side-view CMO diagrams.]{\includegraphics[height=1.6in]{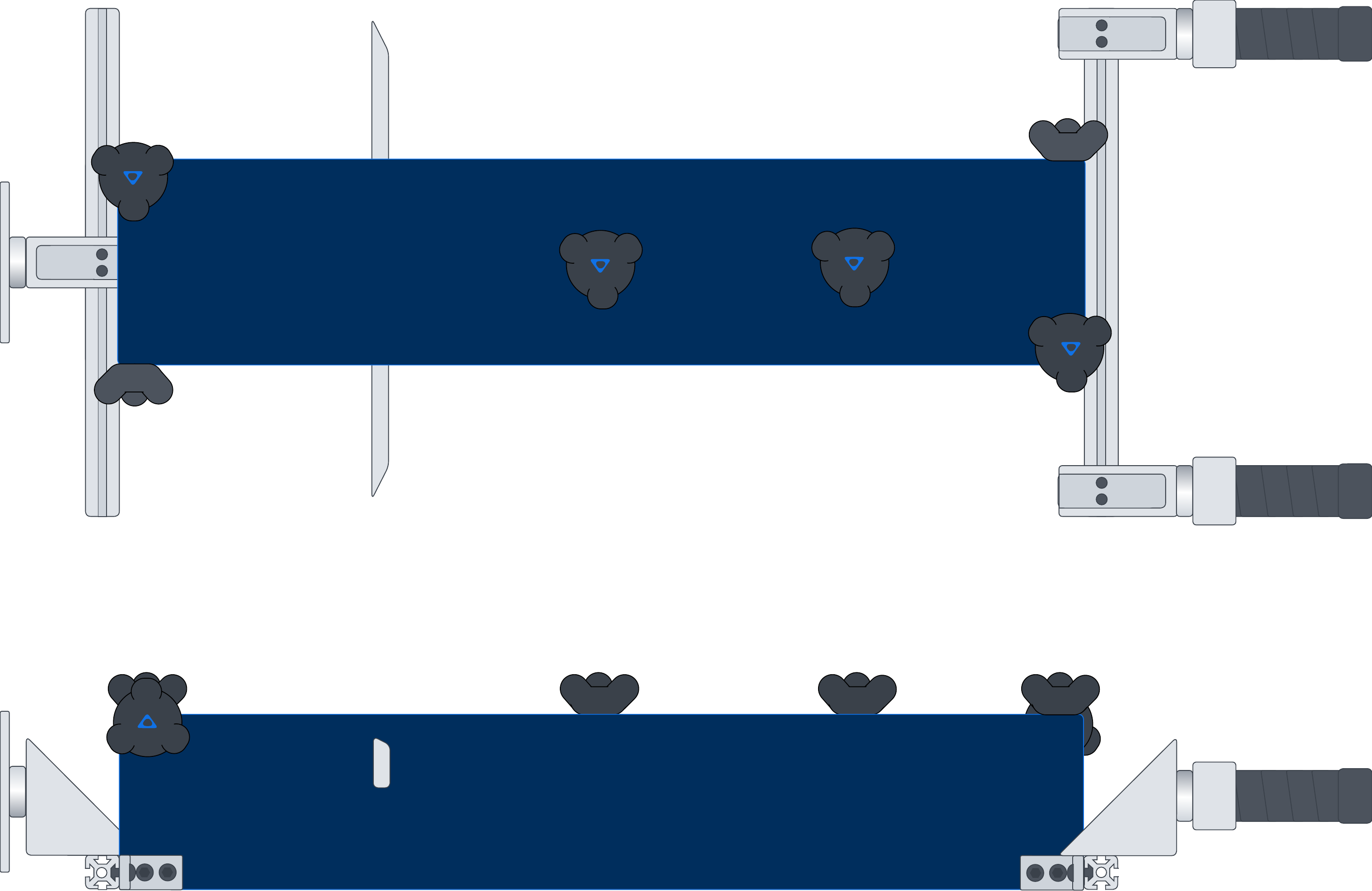}\label{fig:object_topdown_side}\Description{Fully described in the text.}} 
    \hfill 
    \subfigure[Image of the CMO.]{\includegraphics[height=1.6in]{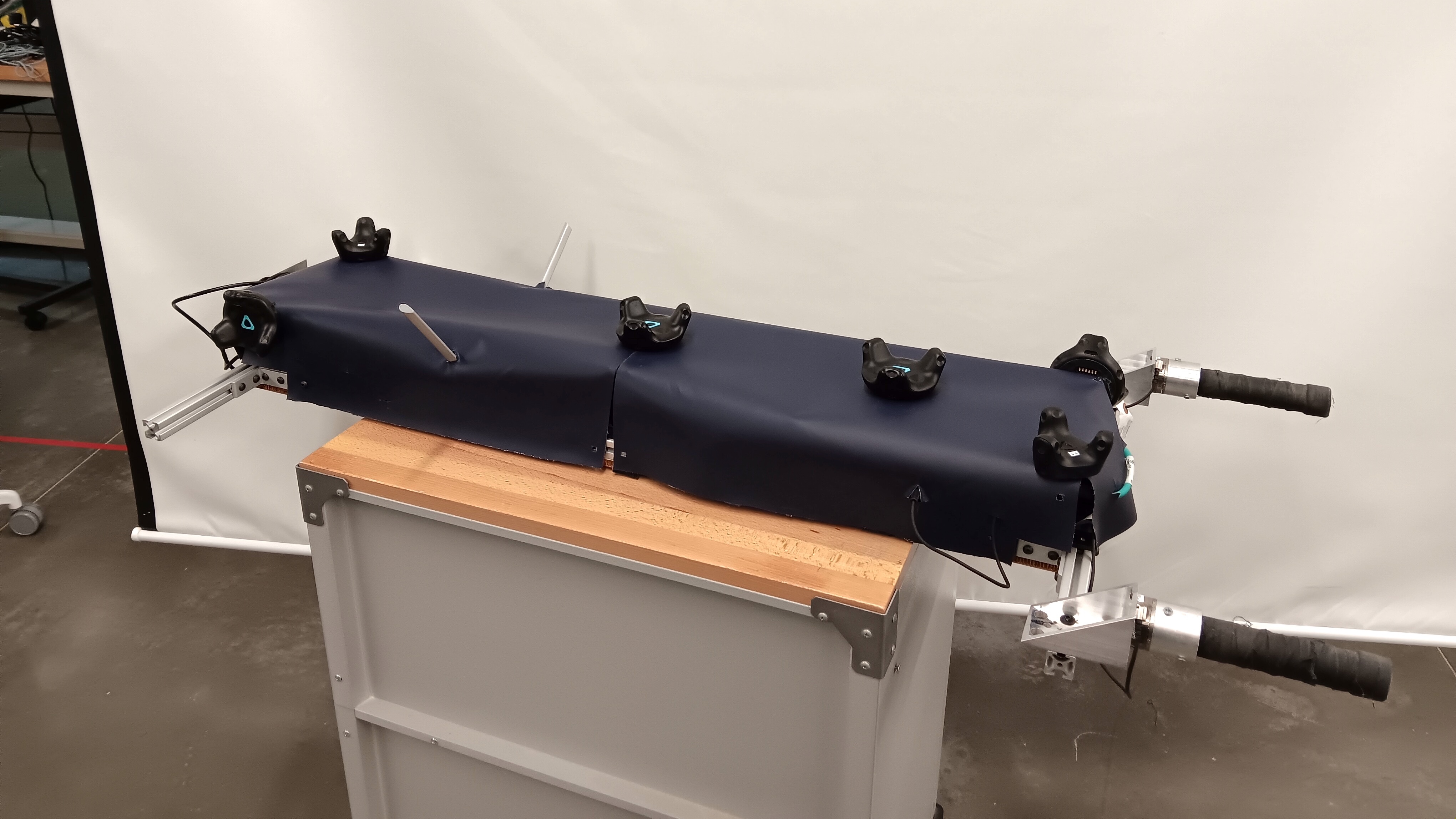}\label{fig:object_real}\Description{Fully described in the text.}}
    \hfill
    \caption{Overview of the co-manipulation object (CMO). The object features an end-effector attachment point (left) and handles for human grasp (right), all mounted to force/torque sensors.}
    \label{fig:object}
    \Description{The object has a rectangular body with handles on one end for the human user and a robotic attachment point on the other. Six triangular black VIVE trackers are attached along the surface of the object, with most trackers distributed symmetrically.}
\end{figure}

The co-manipulation object (CMO) seen in Figure~\ref{fig:object} modifies the design from Shaw et al. \cite{shaw_thesis_2024} and Freeman et al. \cite{freeman_classification_2024}, maintaining the original handle positions but with a centralized robot attachment point and reduced weight (11 kg versus 27 kg). While the long-term goal involves heavy object manipulation, this lighter design prioritizes testing overall functionality. The 1.37 m $\times$ 0.55 m object comprises an extruded aluminum frame on a honeycomb-board panel, two wooden handles for human interaction, and an aluminum plate for the robot interface. Force/torque sensors measure both human and robot inputs, and a system of six VIVE motion trackers serves dual purposes: four trackers capture the robot's kinematic data for post-analysis, while two trackers provide real-time positional data to Unity, enabling precise mapping of virtual objects to their physical counterparts in the VR environment.

    

\subsection{Control}
To ensure proper coordination between the robot arm and mobile base during co-manipulation tasks, two concurrent control strategies are employed. A displacement-based controller maps arm displacement to velocity commands for the mobile base, enabling smooth navigation. Meanwhile, a model reference adaptive controller (MRAC) maintains arm stability by returning it to an initial position. These controllers work together to balance mobility and stability during interactions. The details of these controllers are described in more detail in the following subsections. 


\subsubsection{Displacement-Based Controller}
\label{sec:displacement_control}
Consider a manipulator mounted to a holonomic mobile base whose initial end-effector position, $e$, at time $t=0$ relative to a fixed point on the mobile base, $b$, expressed in the mobile base frame, $mb$, is represented by $p_{b \rightarrow e,0}^{mb}$. Suppose that, at time $t$, a human agent physically displaces the end effector to a position $p_{b \rightarrow e,t}^{mb}$. The incurred displacement can be represented by 

\begin{equation}
	\label{eq:displacement}
	\Delta p_{b \rightarrow e}^{b} = p_{b \rightarrow e,t}^{mb} - p_{b \rightarrow e,0}^{b} .
\end{equation}

This displacement vector is then proportionally mapped to velocity commands to the mobile base

\begin{equation}
	\label{eq:base_velocity}
	v_{mb} = k_p \, \Delta p_{b \rightarrow e}^{mb} .
\end{equation}

In order to keep minor fluctuations of the end effector from commanding velocities to the base, a deadband threshold $\delta_{db}$ is implemented beyond which the end effector must be displaced before any velocity commands are published. This is implemented as
\begin{equation}
    \label{eq:thresholds}
    v_{mb} = 
    \begin{cases} 
      0 & \left| \Delta p_{b \rightarrow e}^{mb} \right| \leq \delta_{db} \\
      k_p \, \left( \left|\Delta p_{b \rightarrow e}^{mb}\right| - \delta_{db} \right) \sgn \left(\Delta p_{b \rightarrow e}^{mb}\right) & \left| \Delta p_{b \rightarrow e}^{mb} \right| > \delta_{db}
   \end{cases}
\end{equation}
where $\left|\Delta p_{b \rightarrow e}^{mb}\right| - \delta_{db}$ establishes the displacement past the deadband threshold and $\sgn \left(\Delta p_{b \rightarrow e}^{mb}\right)$ enforces the appropriate direction.

The proportional nature of this method (such that larger displacements result in greater velocities) lends value to a velocity limiter defined as follows: 

\begin{equation}
    \label{eq:velocity_threh}
    v_{mb} = 
    \begin{cases} 
      v_{mb} & \left| v_{mb} \right| \leq v_{\max} \\
      v_{max} \sgn \left(v_{mb}\right) & \left| v_{mb} \right| > v_{max}
   \end{cases}
\end{equation}

These combine to form Algorithm~\ref{alg:displacement}.

\begin{algorithm}
	\caption{Displacement-based velocity controller.} 
    \label{alg:displacement}
	\begin{algorithmic}[1]
	    \STATE Initialize:  $p_{b \rightarrow e,0}^{mb}$.
	    \WHILE{true}
            \STATE $\Delta p_{b \rightarrow e}^{mb} = p_{b \rightarrow e,t}^{mb} - p_{b \rightarrow e,0}^{mb}$
	        \IF{$\left| \Delta p_{b \rightarrow e}^{mb} \right| \leq \delta_{db}$}
                \STATE $v_{mb} = 0$
            \ELSE
                \STATE $v_{mb} = k_p \, \left( \left|\Delta p_{b \rightarrow e}^{mb}\right| - \delta_{db} \right) \sgn \left(\Delta p_{b \rightarrow e}^{mb}\right)$
            \ENDIF
            \IF{$\left|v_{mb}\right| > v_{max}$}
                \STATE $v_{mb} = v_{max}\sgn\left(v_{mb}\right)$
            \ENDIF
            \STATE Send velocity command $v_{mb}$.
	    \ENDWHILE
	\end{algorithmic}
\end{algorithm}

The control algorithm creates an implicit negative feedback loop: base motion reduces arm deformation, which in turn reduces the commanded velocity, leading to stable behavior. Its simplicity allows straightforward implementation across all three translational axes, provided the physical platform supports it.\footnote{In practice, this algorithm is implemented element-wise, with thresholds and maximum velocity checked individually for each axis. Since our robotic platform does not support vertical motion, any deformation along the vertical axis is ignored.} 
Although similar to admittance control, our approach does not rely on measured forces as the input. Instead, the controller uses end-effector displacement—a direct result of applied forces—to calculate velocity commands. This distinction eliminates the need for force sensing while maintaining functionality comparable to traditional admittance control.

This algorithm relies on the human’s ability to displace the end effector and the robot’s proprioceptive ability to detect that movement.
To achieve end-effector displacement, some form of compliance is necessary. Compliance can be introduced passively or actively. Passive methods \cite{ham_compliant_2009} include series elastic actuators, flexible linkages, and low-torque, back-drivable motors, while active methods typically involve virtual impedance or admittance-based control \cite{calanca_review_2016, abu-dakka_variable_2020}. Our approach employs passive compliance through a pneumatically actuated, soft, continuum-joint robot arm.

By using a passively compliant soft arm, we eliminate the need for force sensing, active force control, or virtual impedance models. Instead, the arm pressures can be tuned to have desirable impedance characteristics. This arm structure not only provides the compliance necessary to operate the controller but also enhances safety and creates a more natural interaction experience.

\begin{figure}
    \centering
    \includegraphics[width=0.6\linewidth]{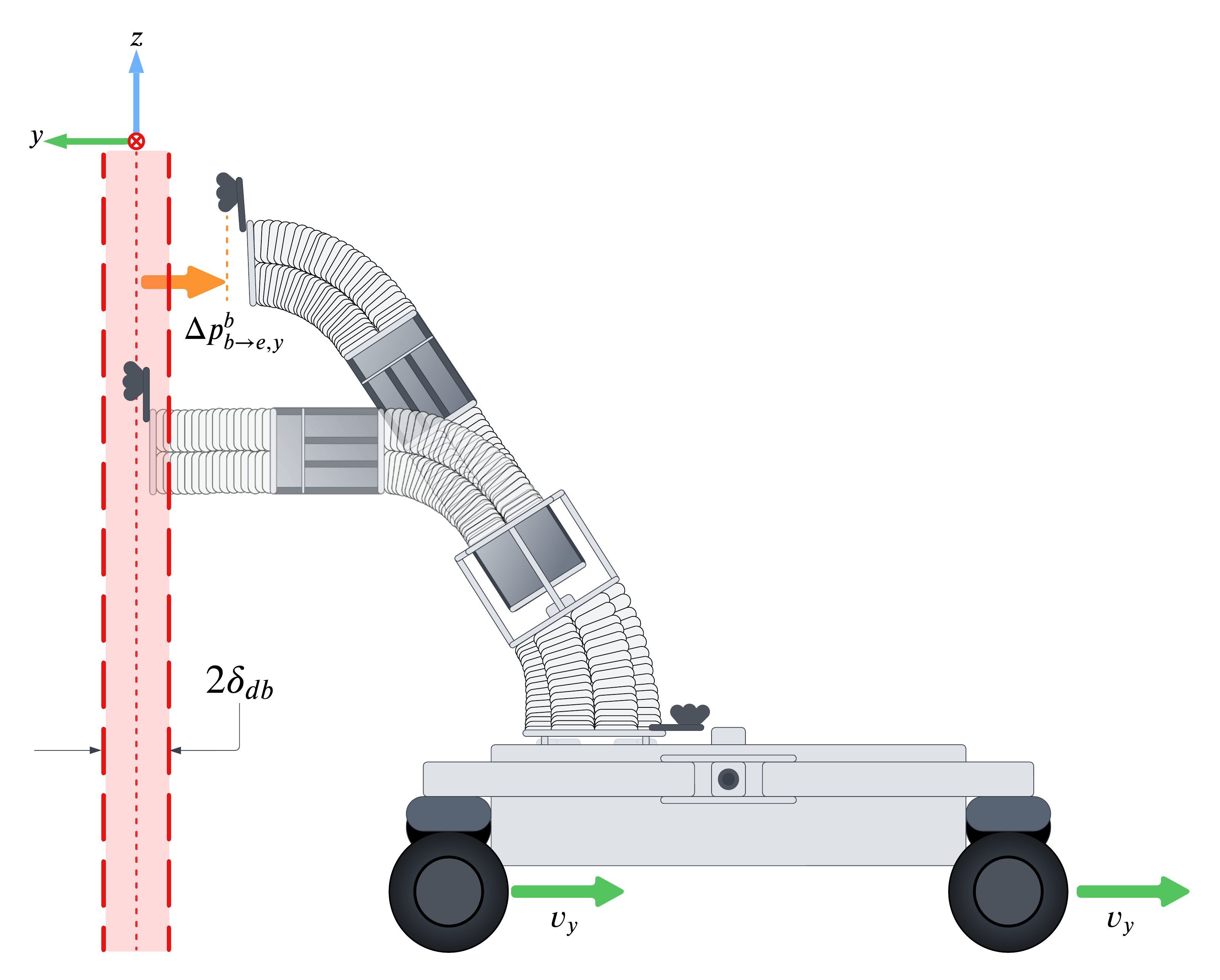}
    \caption{Visualization of displacement-based control algorithm.}
    \label{fig:controller}
    \Description[Side view of the robot showing two arm configurations: one within the deadband region and one deformed beyond it. Wheel velocity is proportional to arm deformation along the same axis.]{Side view of the robot, showing the mobile base and a soft arm. The image includes two configurations of the arm: an initial configuration within the deadband region and a later configuration where the arm is deformed back beyond the deadband threshold. The deadband region is depicted as a shaded area in front of the base, illustrating the range where small displacements do not trigger corrective motion. For the second configuration, the amount of the arm's deformation beyond the deadband threshold is proportional to the velocity commands sent to the wheels, with the velocity occurring along the same axis as the arm deformation.}
\end{figure}

Soft robots traditionally face challenges with precise proprioception. Therefore, in order to gather the data necessary to enact our displacement-based control, VIVE motion trackers are placed on the robot end effector and the mobile base to provide position and orientation data. Furthermore, orientation data gathered from additional VIVE motion trackers placed at the extremities of each joint can be used in tandem with the constant curvature assumption \cite{allen_closed_2020, hyatt_model_2020} to provide accurate configuration estimation. While our soft robot is reliant on these trackers, work is ongoing to improve these assumptions \cite{sorensen_soft_2023}.


This control approach closely resembles that of Stuckler and Behnke \cite{stuckler_following_2011}, who used a displacement-input, velocity-output controller built on an impedance controller for a two-arm rigid robot on a mobile base. Unlike our implementation, their approach calculated rotations by using the relative displacement of each arm to determine an angle. Our approach is also similar to Rauniyar et al. \cite{rauniyar_mewbots_2021}, who utilized a linear potentiometer to measure force-induced displacements to drive their platform, as opposed to our approach which relies on the natural deformation of a soft robot arm.

\subsubsection{MRAC}
\label{sec:mrac}
To ensure proper displacement-based control, the arm must accurately return to its initialized position relative to the mobile base, $p_{b\rightarrow e,0}^{mb}$. While the use of constant pressure control--a lower-level control strategy where the individual chambers in each joint are maintained at a consistent pressure level--performed adequately, it did not ensure that the arm would consistently return to the same position. Instead, it often converged to a local minimum that satisfied the constant pressure parameters, while falling outside the deadband thresholds. To overcome this issue, model reference adaptive control (MRAC) was implemented at the joint level to drive each joint of the unloaded arm to a 30 degree reference angle such that the end effector  would be perpendicular to the ground plane and facing the participant and object (see Figure~\ref{fig:vr_real-space}). 

With the arm attached to the object, the adaptive gain for MRAC would build to account for the added loading until the joint angles reached steady state. Subsequently, the gain was set to zero to prevent the accumulation of integrator-like joint forces that could cause undesirable pushing and pulling as the arm deformed during normal displacement control.




\subsection{Experiment Design}
The experiment consisted of individuals performing co-manipulation tasks with the robotic platform described in Section~\ref{sec:robot-platform}. Tasks required the human participant to move the co-manipulation object (CMO) described in Section~\ref{sec:object} with the robot to a goal position indicated via a virtual reality (VR) headset (see Figure~\ref{fig:vr_real-space}). The movement of the mobile base was determined using displacement-based control, and the pneumatic arm was controlled using Model Reference Adaptive Control (MRAC), both discussed in Sections~\ref{sec:displacement_control} and~\ref{sec:mrac} respectively.

\subsubsection{IRB and Recruitment}
This experiment was performed under the approval of the BYU's Institutional Review Board (IRB) to ensure ethical research was performed and that risks were limited and any risks were transparently communicated to participants. Recruitment consisted of in-person advertising, personal invitations, fliers, advertisements on displays in the BYU Engineering Building, and visits to classes taught by BYU engineering professors. Participants ranged in age from 18-60, with an average age of 24. 69\% of participants were male and the remaining 31\% were female (self-reported in all cases). Other demographic information was not collected, though, since public advertisement was limited to the BYU College of Engineering it is likely that most of the participants were students of the college.

\subsubsection{Virtual Reality System}
\label{sub:vr-system}

\begin{figure}
    \hfill
    \subfigure[Participant performing a co-manipulation task while wearing a VR headset.]{\includegraphics[height=2.3in]{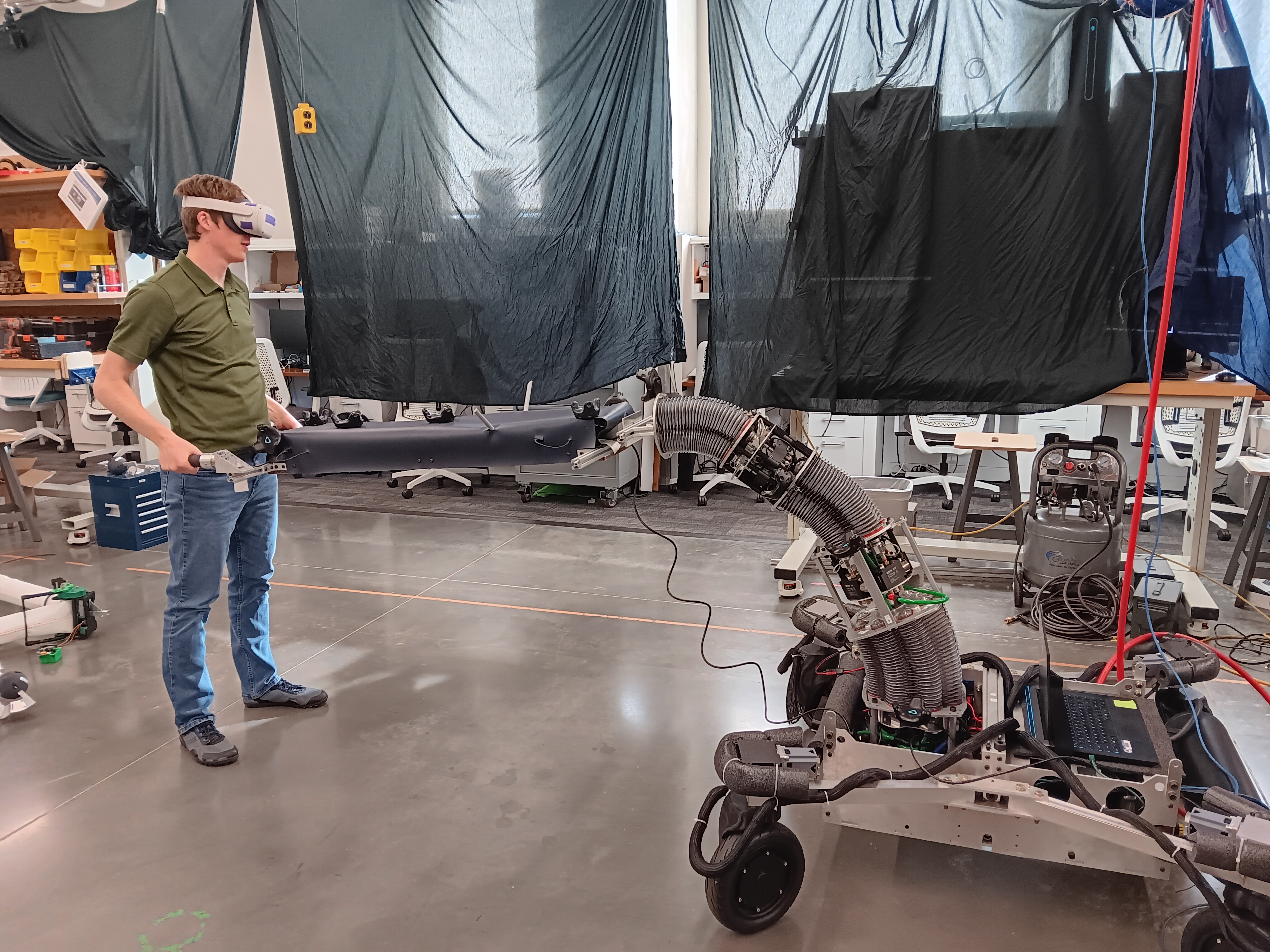}\label{fig:vr_real-space}\Description{A participant wearing a VR headset holds aloft one end of the co-manipulated object, while the robot holds up the other end. The participant, robot, and object are positioned in a workspace cleared of any obstacles.}} 
    \hfill 
    \subfigure[Participant view in the virtual environment.]{\includegraphics[height=2.3in]{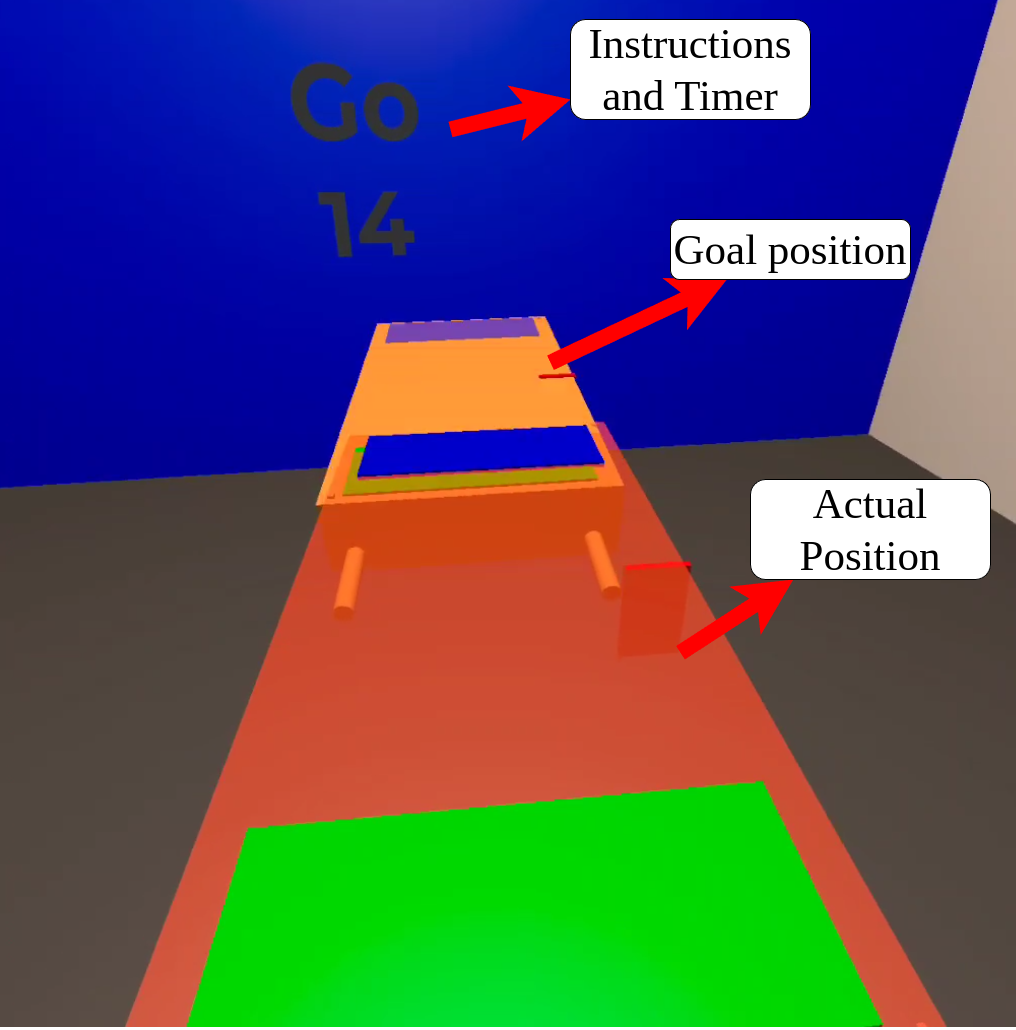}\label{fig:vr_virtual-space}\Description{Fully described in the text.}}
    \hfill
    \caption{Overview of the virtual environment setup for the co-manipulation tasks.}
    \label{fig:vr-and-real-spaces}
    \Description{Fully described in the text.}
\end{figure}

The VR system used in this study was built in Unity, a popular game development software that is well integrated with virtual reality systems such as Oculus (now MetaQuest) and VIVE. The Unity game manages the task order and timing, and coordinates between the VIVE and Oculus systems by receiving motion tracking data from VIVE, updating the VR environment with that data, and broadcasting the VR environment to the Oculus headset. The VIVE system is used for motion tracking of the co-manipulation object (CMO), data collection, and to align the Oculus virtual environment with the physical environment. The Oculus is also used to instruct participants in how to move. An example image of the Oculus display can be found in Figure~\ref{fig:vr_virtual-space}. In VR, the participant sees two representations of the CMO, one opaque and the other transparent. The transparent CMO stays aligned with the actual CMO, while the opaque CMO indicates the goal position. The goal CMO moves to a location within the environment and the participant must align the transparent CMO with the goal to accomplish the task. The Oculus also shows the participant some minor instructions such as the word ``go'' and a timer counting the elapsed time since the task's beginning. For safety, the physical space is cleared of obstacles and the Oculus play area is set such that if the participant gets too close to the edge of the physical space the Oculus displays the physical space instead of the virtual space so the participant can see where they are and know that they have left the VR area.

\subsubsection{Study Procedure}
A standardized script guided participants through the experiment, ensuring consistent instructions across all participants. At the start, participants were introduced to their robot partner, given an overview of the study procedure, and asked to fill out audio, video, and photography release forms. Participants were then presented with a pre-study survey gauging their impressions of the robot as a partner. 


Before entering the VR environment, participants were allowed up to five minutes to familiarize themselves with the robot by collaboratively moving the CMO around the experimental workspace. This served to reduce the effects of learning in our experiment and encourage more consistent performance across participants regardless of past experience with VR environments. Furthermore, it provided the opportunity to address any safety concerns that participants had and provide a baseline of normal operation.

After the initial familiarization, participants entered the VR environment and were positioned in front of the co-manipulation object (CMO). Participants were informed they would be timed, emphasizing the importance of task completion speed. After grasping the table handles, the support holding the CMO was removed, and the experiment would start.

\begin{figure}
    \hfill
    \subfigure[All eight motion tasks participants perform during a set. Each double-ended arrow represents two tasks, a motion in one direction and its inverse.]{\includegraphics[height=1.6in]{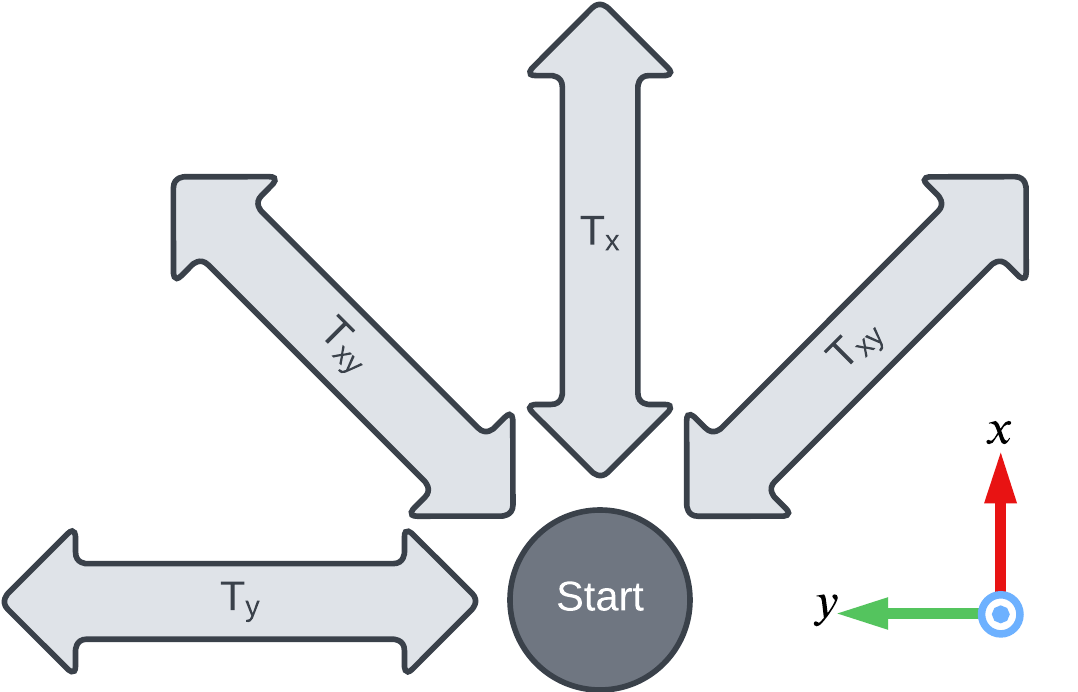}\label{fig:tasks}\Description[A node with four labeled arrows showing translations in x, y, and both diagonals.]{A single node with four double-sided arrows extending: one to the left labeled 'T sub y' (translation in y), one diagonally left and up labeled 'T sub x y' (translation in both x and y), one straight up labeled 'T sub x' (translation in x), and one diagonally right and up labeled 'T sub x y' (translation in both x and y). The coordinate frame defines up as a positive translation in x and left as a positive translation in y.}} 
    \hfill 
    \subfigure[An illustration of all possible task combinations constrained so participants would stay with in a $2 m \times 2 m$ region, centered around the start point.]{\includegraphics[height=2.8in]{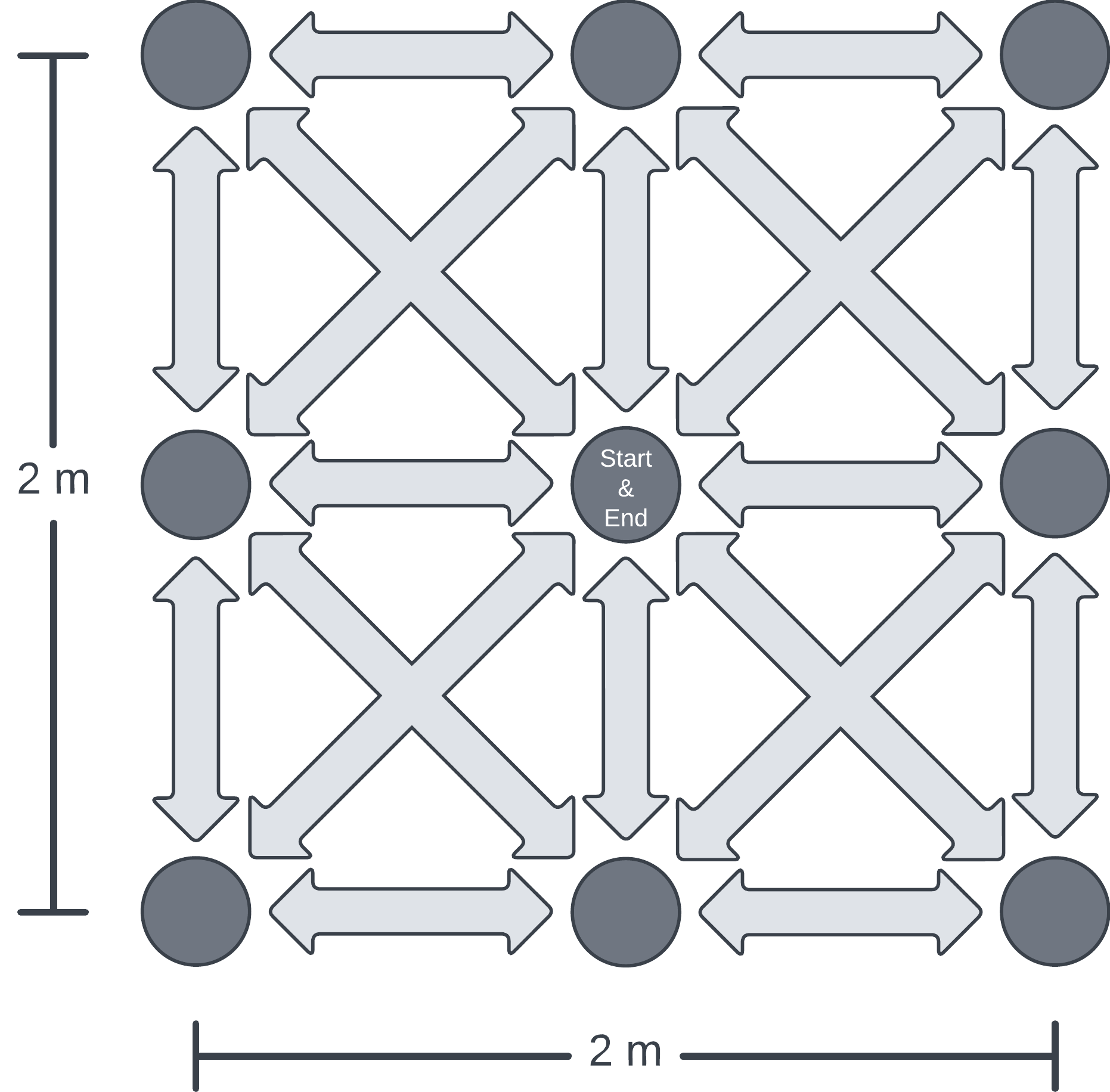}\label{fig:task_space}\Description[A 3x3 grid with nine nodes connected by double-sided arrows to their neighbors. The center node is labeled as the start and end.]{A 2-meter by 2-meter grid with nine nodes arranged in three rows and three columns. Double-sided arrows connect each node to its immediate neighbors. The center node is labeled as the start and end point.}}
    \hfill
    \caption{Graphical representations of (a) the various tasks and (b) the task space.}
    \label{fig:task-visuals}
    \Description[]{Fully described in the text.}
\end{figure}

Given the difficulty of distinguishing between rotation and translation-based tasks (see\cite{jensen2021trends, mielke2017analysis, Arai2000, Dumora2012}), this experiment focuses on eight planar, translational tasks: positive and negative 1 meter translations in forward/backward ($x$) and left/right($y$) directions, and four diagonal combinations of $x$ and $y$ motions. These tasks are represented in Fig. \ref{fig:tasks} and are a subset of the tasks performed by Shaw \cite{shaw_thesis_2024}. The completion of all 8 tasks represents a set. The order of the presented tasks were randomized for each set. To account for physical space limitations, the number of possible task combinations (40,320) were filtered to only consider those that would fit within a $2 m \times 2 m$ region, making 5,664 unique sets (see Fig. \ref{fig:task_space}). A video of a participant completing a single set is included as supplemental material and can be found at \hyperlink{https://youtu.be/4HnFI5s7RrQ}{https://youtu.be/4HnFI5s7RrQ}.

Upon starting the experiment, participants would perform all eight tasks of a randomly selected set, with short five-second pauses between each task. Though encouraged to work quickly, participants were given as much time as needed to successfully complete each task. Due to the symmetric nature of the tasks, participants would start and end each set in the center of the experimental workspace. When participants completed the set, a researcher would replace the support under the CMO, leaving the participant free to remove their VR headset and begin the NASA TLX workload assessment associated with the completed set.



\begin{figure}[tb]
    \centering
    \includegraphics[width=\linewidth]{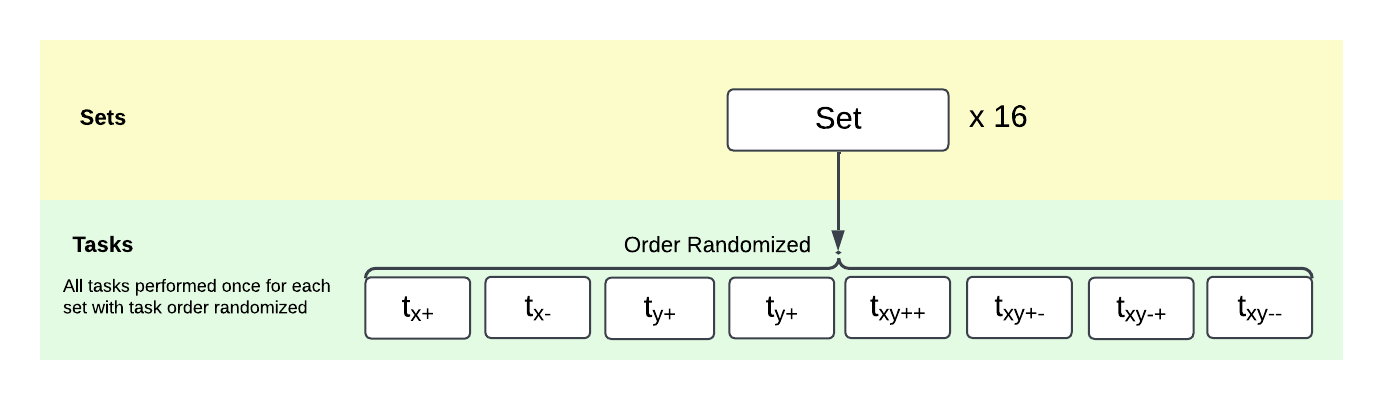}
    \caption{Diagram of the process for a study session. All eight tasks (green) were performed in a randomized order for every set (yellow). Each participant performed 16 sets, from which 6 were used for analysis in this work.}
    \label{fig:hr-process-flowblock}
    \Description[A "set" box (with "x16" to its right) has a downward arrow leading to a row of eight boxes grouped by a curly brace. Each box is labeled with a planar translation.]{A flowchart beginning with a rectangular box on a yellow background containing the word "set." To the right of this box, the label "x16" indicates that the "set" operation is repeated sixteen times. From the bottom of the "set" box, a downward-pointing arrow leads to a horizontal row of eight smaller boxes on a green background. A curly brace, whose pointed end connects with the arrow, arches over these eight boxes like an umbrella, visually grouping them together. Each of the eight boxes contains a translation command: individual positive and negative translations in x and y, a positive translation in both x and y, a positive translation in x and a negative translation in y, a negative translation in x and a positive translation in y, and a negative translation in both x and y.}
\end{figure}

Participants performed a total of 16 sets of co-manipulation tasks with the robot before the study was completed. Beyond serving as a benchmark comparison to human-human co-manipulation experiments, this study likewise modified aspects of the robot's behavior between each set to determine optimal behaviors associated with the manipulator's relative stiffness as well as the spatial reference point used for the displacement-based controller.  The results of these findings are beyond the scope of this work. For quantitative comparisons to human-human co-manipulation studies, we will compare the data from six of the 16 sets of this study that had consistent parameters (i.e., the center of the CMO acting as the spatial reference point for the displacement-based controller and a constant stiffness setting for the robot arm). A flowchart of a typical study session is included in Figure~\ref{fig:hr-process-flowblock}.



Once the participants had completed all sets in the study, they completed a post-study survey regarding their experience and their perceptions of the robot partner. 

The pre- and post-study survey questions were modeled on those developed by Freeman \cite{freeman_motion_2022}, a subset of which were used by \cite{shaw_thesis_2024}. Using the same questions ensured comparability of our results to both Freeman and Shaw's human-human studies. Within the surveys, participants answer questions ranking metrics on a semantic differential scale (e.g., ``Please rate your partner on a scale from one to seven: 1 = Inexperienced, 7 = Experienced'') or else on a 5-point Likert scale (e.g., ``Qualify how well the interaction with your partner went. Please answer the questions on a scale from `Not at all' to `Very.' `How responsive was your partner to the motions that you made?' ''). Metrics include concepts such as partner trustworthiness, safety, perceived control, comfort, and reliability.

\section{Results}
\label{sec:results}

Our analysis incorporates both quantitative and qualitative metrics, which we compare to Shaw et al.'s human-human study findings \cite{shaw_thesis_2024}. We also draw qualitative comparisons with Freeman et al.'s work \cite{freeman_classification_2024}. Though these studies employed different specific tasks, their similar research methodologies and co-manipulation approaches enable meaningful comparison when accounting for contextual differences.

Shaw studied motion primitives of human-human dyads using VR headsets, where leaders could see both the object and the goal while followers saw only the object. Shaw's motion primitives closely matched our own, but with a more extensive range of motions, making his work a valuable comparative reference. Among Shaw's various configurations (leader-leader, leader-follower, and leader-follower-follower), we focus on the leader-follower setup as it most closely matches our framework with a robotic-follower seeking to respond to a human-leader. Freeman et al. likewise investigated co-manipulation in human pairs, though they navigated obstacle-filled environments, with both participants acting as informed leaders, making the motion comparison less direct. However, the co-manipulation element with a human-partner makes his work excellent for qualitative comparison.

\subsection{Quantitative Performance Comparison}
To quantitatively compare human-human and human-soft-robot performance, we analyze completion time, scaled path length, and velocity magnitude distributions. The scaled path length normalizes the total path length traversed in a task by dividing that path length by the direct distance between the start and end points. This means that a scaled path length of 1.0 indicates the shortest possible path. This scaling enables comparison between diagonal tasks (1.41 m) and straight tasks (1.0 m), while accounting for variations in starting positions. Our quantitative comparison includes only tasks common to this study and Shaw's \cite{shaw_thesis_2024}. Tasks involving vertical translation or rotation from the human-human study are excluded, as are the $t_{xy+-}$ and $t_{xy-+}$ tasks which were unique to the human-soft-robot study. We refer to Shaw's human-human data as HH and the human-soft-robot data in this paper as HSR throughout this section.

    
    

\subsubsection{Completion Times}

The HH completion times were consistently lower than HSR completion times, averaging 11.9\% (0.72 s) shorter across all tasks. A detailed breakdown is provided in Table~\ref{tab:hh-b-completion-time}. When considering all tasks, this difference was statistically significant under a Bonferroni-corrected Brunner-Munzel test. Three of the six individual tasks also showed statistically significant differences, with human-human teams consistently achieving shorter completion times than human-soft-robot teams.



\begin{table}[ht]
\centering
\caption{Human-Human vs Human-Robot task completion times with mean and standard deviation, as well as the p-value resulting from a Bonferroni-Corrected Brunner-Munzel test of significant differences between the HH and HSR data for each task. P-values less than 0.05 are in bold for emphasis.}
\begin{tabular}{lccc}
\toprule
Task & HH Completion Time & HSR Completion Time & P-value \\
     & (Mean ± SD) & (Mean ± SD)  &  \\
\midrule
$t_{x+}$      & 4.69 $\pm$ 1.70 & 4.91 $\pm$ 1.84 & 0.380 \\
$t_{x-}$      & 4.10 $\pm$ 1.58 & 5.19 $\pm$ 1.86 & \textbf{\(<\)0.0001} \\
$\mathbf{t_{x\pm}}$ & 4.38 $\pm$ 1.67  & 5.05 $\pm$ 1.85 & \textbf{0.0004} \\
$t_{y+}$      & 5.31 $\pm$ 2.40  & 6.07 $\pm$ 2.15 & \textbf{0.009} \\
$t_{y-}$      & 5.53 $\pm$ 2.44  & 6.05 $\pm$ 2.12 & 0.063\\
$\mathbf{t_{y\pm}}$ & 5.43 $\pm$ 2.42 & 6.06 $\pm$ 2.14 & \textbf{0.002} \\
$t_{xy++}$    & 6.31 $\pm$ 1.92 & 7.18 $\pm$ 2.66 & 0.054 \\
$t_{xy--}$    & 6.03 $\pm$ 1.68 & 6.82 $\pm$ 2.26 & \textbf{0.048} \\
$\mathbf{t_{xy\pm\pm}}$ & 6.17 $\pm$ 1.81 & 7.00 $\pm$ 2.49 & \textbf{\(<\)0.0001} \\
\cmidrule(lr){1-4}
\textbf{Overall} & 5.32 $\pm$ 2.12 & 6.04 $\pm$ 2.30 & \textbf{\(<\) 0.0001} \\
\bottomrule
\label{tab:hh-b-completion-time}
\end{tabular}
\end{table}

\subsubsection{Scaled Path Lengths}
The scaled path lengths for tasks in the HH data are lower than those in the HSR data for all tasks except for $t_{x+}$. On average, the HH scaled path length is 1.21 and the HSR scaled path length is 1.26. This overall difference is statistically significant, but it is noteworthy that the difference in scaled path lengths for individual task types was only statistically significant for $t_{x-}$ tasks. The low difference between scaled path lengths indicates very similar performance. As such we can conclude that improving the ability of the robot to follow a more efficient path is a lower priority than improving the ability of the human-soft-robot teams to match human-human completion time. A more in-depth breakdown of results for scaled path length for each task can be found in Table~\ref{tab:hh-b-scaled-path-length}.

\begin{table}[ht]
\centering
\caption{Human-Human vs Setting B Scaled Path Lengths}
\begin{tabular}{lccc}
\toprule
Task & HH Scaled Path Length & HSR Scaled Path Length & P-value \\
     & (Mean ± SD) & (Mean ± SD)  &  \\
\midrule
$t_{x+}$      & 1.14 $\pm$ 0.16 & 1.11 $\pm$ 0.18 & 0.765 \\
$t_{x-}$      & 1.10 $\pm$ 0.13 & 1.13 $\pm$ 0.23 & \textbf{0.022} \\
$\mathbf{t_{x\pm}}$ & 1.12 $\pm$ 0.15 & 1.12 $\pm$ 0.21 & 0.069\\
$t_{y+}$      & 1.27 $\pm$ 0.34 & 1.30 $\pm$ 0.37 & 0.351 \\
$t_{y-}$      & 1.25 $\pm$ 0.27 & 1.37 $\pm$ 0.48 & 0.137 \\
$\mathbf{t_{y\pm}}$ & 1.26 $\pm$ 0.31 & 1.33 $\pm$ 0.43 & 0.101 \\
$t_{xy++}$    & 1.28 $\pm$ 0.23 & 1.36 $\pm$ 0.33 & 0.151 \\
$t_{xy--}$    & 1.24 $\pm$ 0.24 & 1.25 $\pm$ 0.30 & 0.822 \\
$\mathbf{t_{xy\pm\pm}}$ & 1.26 $\pm$ 0.24 & 1.31 $\pm$ 0.32 & 0.285 \\
\cmidrule(lr){1-4}
\textbf{Overall} & 1.21 $\pm$ 0.25 & 1.26 $\pm$ 0.34 & \textbf{0.049} \\
\bottomrule
\label{tab:hh-b-scaled-path-length}
\end{tabular}
\end{table}

\subsubsection{Velocity Histograms}

\begin{figure}[htbp]
    \hfill
    \subfigure[HH study.]{\includegraphics[width=0.49\textwidth]{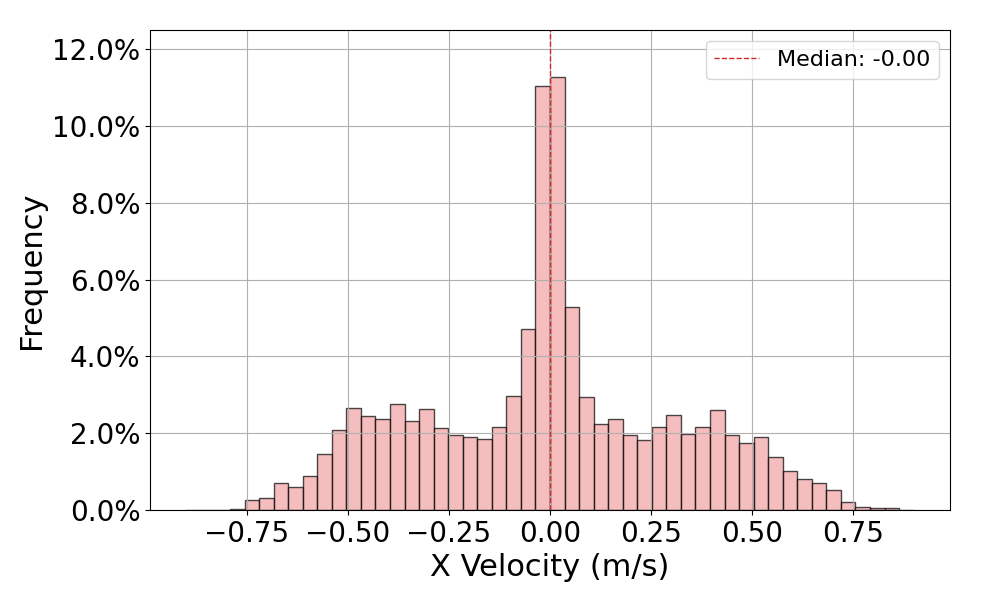}\label{fig:hh-vel-hist}} 
    \hfill 
    \subfigure[HSR study.]{\includegraphics[width=0.49\textwidth]{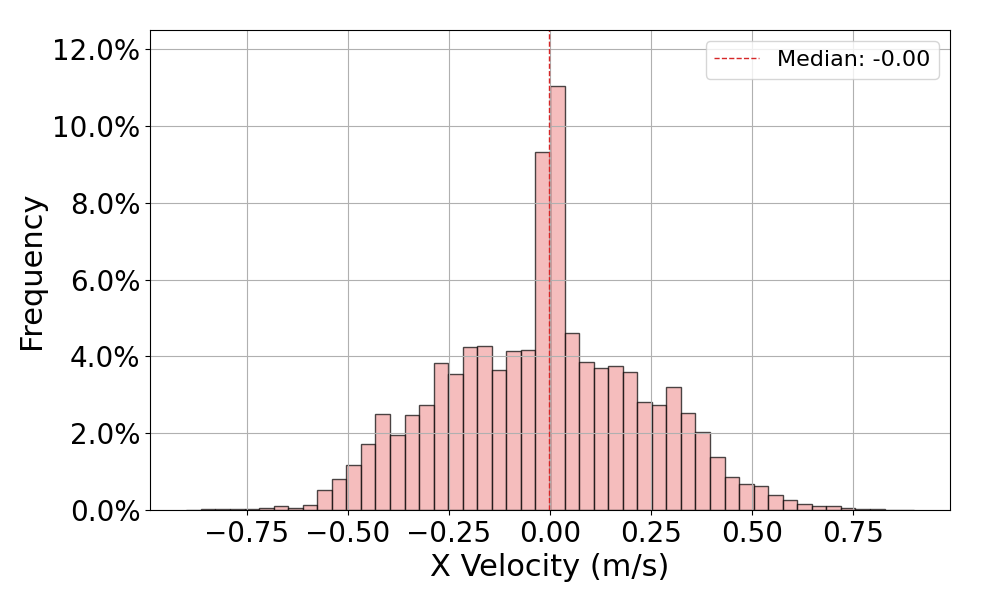}\label{fig:hr-vel-hist}}
    \hfill
    \caption{Histograms of the x-direction velocity for all x-direction tasks for the HH study and the HSR study.}
    \label{fig:hh-b-vel-histograms}
    \Description[]{}
\end{figure}

Analysis of x-direction velocity signals (world frame) for x-direction tasks reveals similar mean velocities near zero for both human-human (HH) and human-soft-robot (HSR) interactions (HH: -0.006 ± 0.32 m/s; HR: -0.003 ± 0.24 m/s) implying that task velocities were symmetrical (Figure~\ref{fig:hh-b-vel-histograms}). However, the histograms show participants in HH trials maintained higher speeds and spent less time stationary compared to HSR trials. This suggests future robot development should focus on enabling higher-speed movements, either by reducing the effort required to achieve these speeds or by improving motion smoothness. A velocity-based controller, operating directly from CMO velocity rather than displacement-derived forces, might better facilitate these faster movements.

\subsection{Qualitative Performance Comparison}

While Freeman et al. \cite{freeman_classification_2024} collected both pre- and post-study responses using identical questions to ours, Shaw \cite{shaw_thesis_2024} employed only post-study responses for a subset of these questions. Given the breadth of survey responses collected, we focus here on the most illustrative findings that highlight key patterns in participant perceptions. For a comprehensive analysis of all survey questions and responses, readers are directed to \cite{cordon_thesis_2024}. 

In examining these human-soft-robot interaction patterns, Shaw's experimental framework provides a particularly relevant point of comparison, as their leader-follower configuration closely mirrors our study's design. While Freeman's work offers valuable insights into human-human co-manipulation, their study's equal-leader structure and full visual access for both participants creates fundamental differences in the interaction dynamic. In contrast, Shaw's asymmetric information setup, where leaders had full environmental awareness while followers operated with limited visibility, better reflects the inherent asymmetry present in our human-soft-robot interactions and should be weighted more heavily.

\begin{table}[]
\caption{Semantic differential scale survey results evaluating human and robot partners. Data includes pre- and post-study responses from Freeman et al. \cite{freeman_classification_2024} (HH-F), post-study responses from Shaw \cite{shaw_thesis_2024} (HH-S, converted from 5-point to 7-point scale), and our human-soft-robot (HSR) study responses.}
\label{tab:qual-pre-post}
\begin{tabular}{@{}rcccc@{}}
\toprule
Metric                                                                    & Condition & Pre-Study       & Post-Study      & Avg. Change      \\ \midrule
                                                                          & HH-F      & 5.75 $\pm$ 1.27 & 6.76 $\pm$ 0.66 & 1.02 $\pm$ 1.39  \\
\multirow{-2}{*}{Untrustworthy (1) - Trustworthy (7)}                     & HSR        & 4.11 $\pm$ 0.99 & 4.47 $\pm$ 1.12 & 0.37 $\pm$ 1.38  \\
\rowcolor[HTML]{EFEFEF} 
\cellcolor[HTML]{EFEFEF}                                                  & HH-F      & 6.10 $\pm$ 1.13 & 6.74 $\pm$ 0.56 & 0.64 $\pm$ 1.14  \\
\rowcolor[HTML]{EFEFEF} 
\multirow{-2}{*}{\cellcolor[HTML]{EFEFEF}Dangerous (1) - Safe (7)}        & HSR        & 4.26 $\pm$ 1.10 & 5.47 $\pm$ 1.12 & 1.21 $\pm$ 1.23  \\
                                                                          & HH-F      & 5.87 $\pm$ 1.18 & 6.56 $\pm$ 1.20 & 0.70 $\pm$ 1.56  \\
\multirow{-2}{*}{Unreliable (1) - Reliable (7)}                           & HSR        & 4.21 $\pm$ 0.79 & 4.36 $\pm$ 1.57 & 0.16 $\pm$ 1.46  \\
\rowcolor[HTML]{EFEFEF} 
\cellcolor[HTML]{EFEFEF}                                                  & HH-F      & 5.48 $\pm$ 1.24 & 6.39 $\pm$ 0.97 & 0.92 $\pm$ 1.39  \\
\rowcolor[HTML]{EFEFEF} 
\multirow{-2}{*}{\cellcolor[HTML]{EFEFEF}Careless (1) - Careful (7)}      & HSR        & 4.21 $\pm$ 0.85 & 4.79 $\pm$ 1.23 & 0.58 $\pm$ 1.50  \\
                                                                          & HH-F      & 5.43 $\pm$ 1.17 & 6.30 $\pm$ 1.10 & 0.86 $\pm$ 1.25  \\
                                                                          & HH-S      & ---             & 5.09 $\pm$ 1.50 & ---              \\
\multirow{-3}{*}{Inexperienced (1) - Experienced (7)}                     & HSR        & 3.79 $\pm$ 0.92 & 3.89 $\pm$ 1.10 & 0.11 $\pm$ 1.10  \\
\rowcolor[HTML]{EFEFEF} 
\cellcolor[HTML]{EFEFEF}                                                  & HH-F      & 5.94 $\pm$ 1.31 & 6.61 $\pm$ 0.91 & 0.67 $\pm$ 1.45  \\
\rowcolor[HTML]{EFEFEF} 
\cellcolor[HTML]{EFEFEF}                                                  & HH-S      & ---             & 5.56 $\pm$ 1.67 & ---              \\
\rowcolor[HTML]{EFEFEF} 
\multirow{-3}{*}{\cellcolor[HTML]{EFEFEF}Unqualified (1) - Qualified (7)} & HSR        & 4.31 $\pm$ 1.16 & 4.37 $\pm$ 1.38 & 0.05 $\pm$ 2.07  \\
                                                                          & HH-F      & 5.56 $\pm$ 1.29 & 6.52 $\pm$ 0.95 & 0.96 $\pm$ 1.51  \\
                                                                          & HH-S      & ---             & 5.25 $\pm$ 1.43 & ---              \\
\multirow{-3}{*}{Unskilled (1) - Skilled (7)}                             & HSR        & 3.84 $\pm$ 0.69 & 4.32 $\pm$ 1.20 & 0.47 $\pm$ 1.47  \\
\rowcolor[HTML]{EFEFEF} 
\cellcolor[HTML]{EFEFEF}                                                  & HH-F      & 6.30 $\pm$ 0.87 & 6.59 $\pm$ 1.12 & 0.29 $\pm$ 1.08  \\
\rowcolor[HTML]{EFEFEF} 
\cellcolor[HTML]{EFEFEF}                                                  & HH-S      & ---             & 5.50 $\pm$ 1.40 & ---              \\
\rowcolor[HTML]{EFEFEF} 
\multirow{-3}{*}{\cellcolor[HTML]{EFEFEF}Incompetent (1) - Competent (7)} & HSR        & 4.47 $\pm$ 1.02 & 4.47 $\pm$ 1.22 & 0.00 $\pm$ 1.67  \\
                                                                          & HH-F      & 6.24 $\pm$ 0.92 & 6.65 $\pm$ 0.77 & 0.39 $\pm$ 0.92  \\
\multirow{-2}{*}{Unintelligent (1) - Intelligent (7)}                     & HSR        & 4.26 $\pm$ 1.24 & 4.05 $\pm$ 1.43 & -0.21 $\pm$ 1.96 \\
\rowcolor[HTML]{EFEFEF} 
\cellcolor[HTML]{EFEFEF}                                                  & HH-F      & 5.15 $\pm$ 1.12 & 5.89 $\pm$ 1.19 & 0.75 $\pm$ 1.38  \\
\rowcolor[HTML]{EFEFEF} 
\multirow{-2}{*}{\cellcolor[HTML]{EFEFEF}Inexpert (1) - Expert (7)}       & HSR        & 3.84 $\pm$ 0.69 & 3.74 $\pm$ 1.24 & -0.11 $\pm$ 1.20 \\ \bottomrule
\end{tabular}
\end{table}

Pre-study responses showed neutral attitudes toward robot partners, contrasting with the strongly positive expectations for human partners (see Table~\ref{tab:qual-pre-post}). This disparity may stem from participants' natural comfort with human interaction versus uncertainty about working with an unfamiliar, non-anthropomorphic robot. A more human-like robot design could influence these initial perceptions.

\begin{figure}[htbp]
    \hfill
    \subfigure[I felt I could rely on my partner to do what they were supposed to.]{\includegraphics[width=0.49\textwidth]{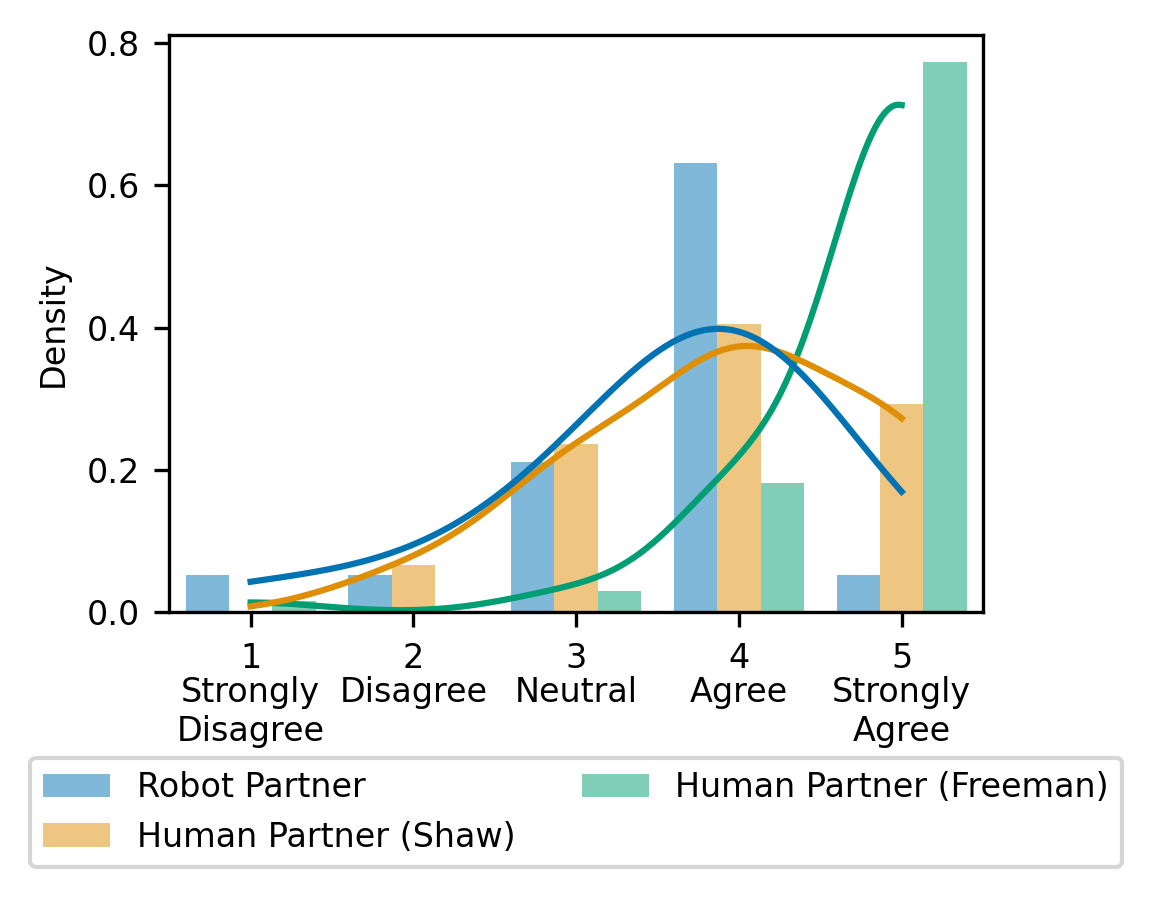}\label{fig:qual-rely}}
    \hfill
    \subfigure[The interaction with my partner flowed smoothly.]{\includegraphics[width=0.49\textwidth]{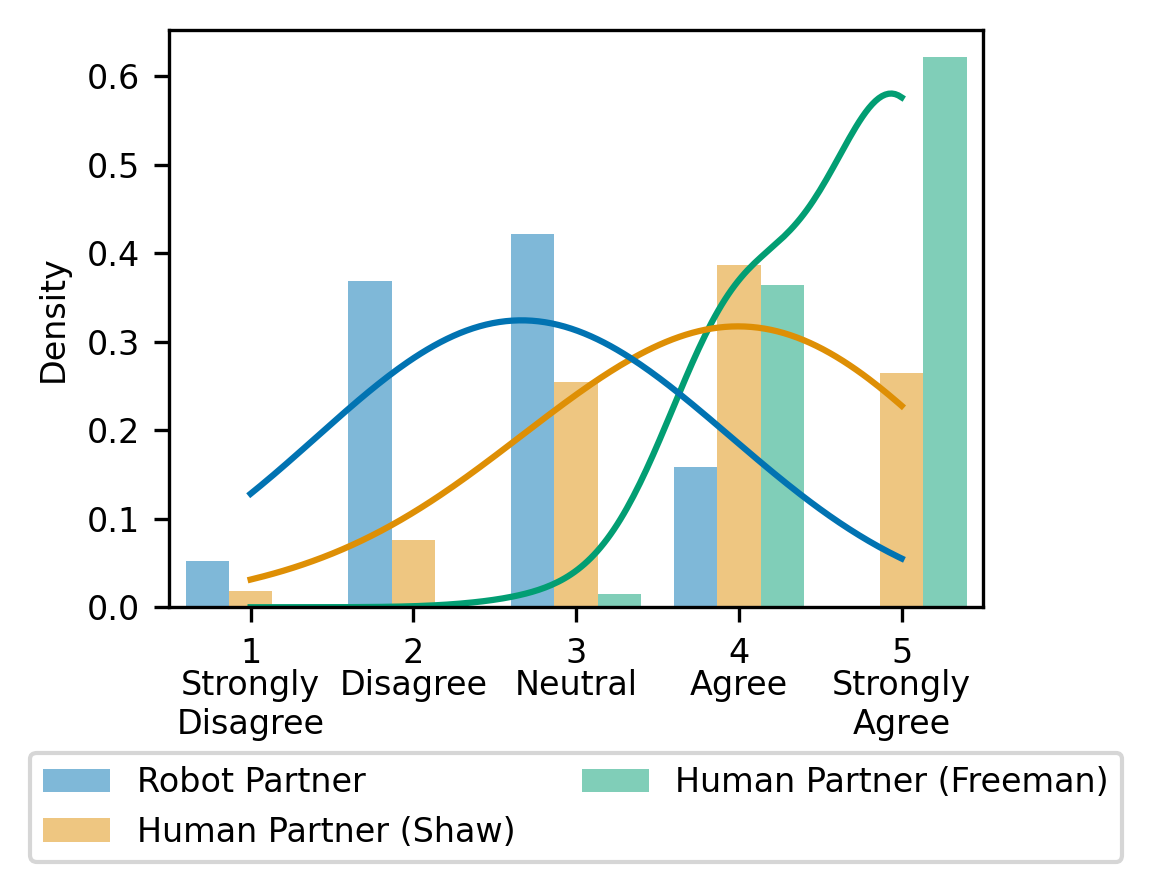}\label{fig:qual-smooth}} 
    \subfigure[We each got to move how we wanted.]{\includegraphics[width=0.49\textwidth]{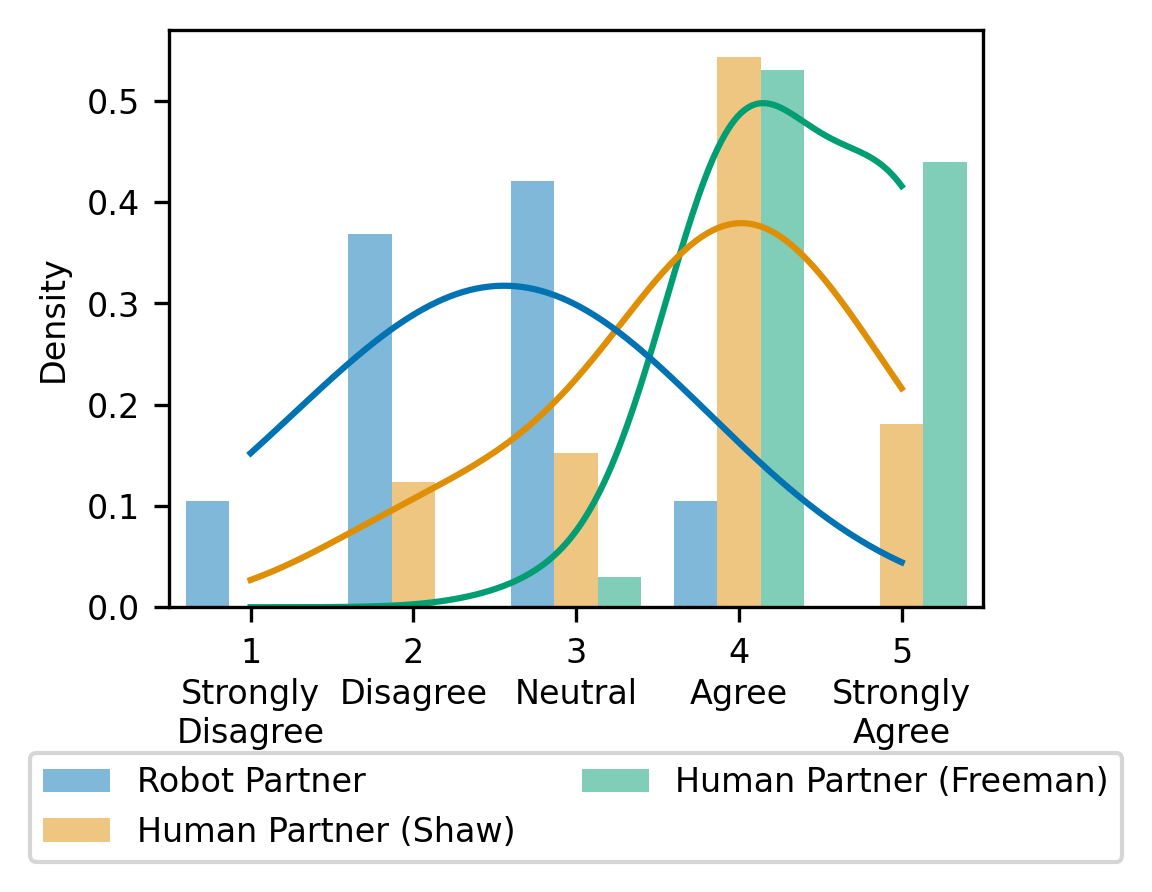}\label{fig:qual-move}}
    \hfill
    \subfigure[My partner went slower than I wanted to.]{\includegraphics[width=0.49\textwidth]{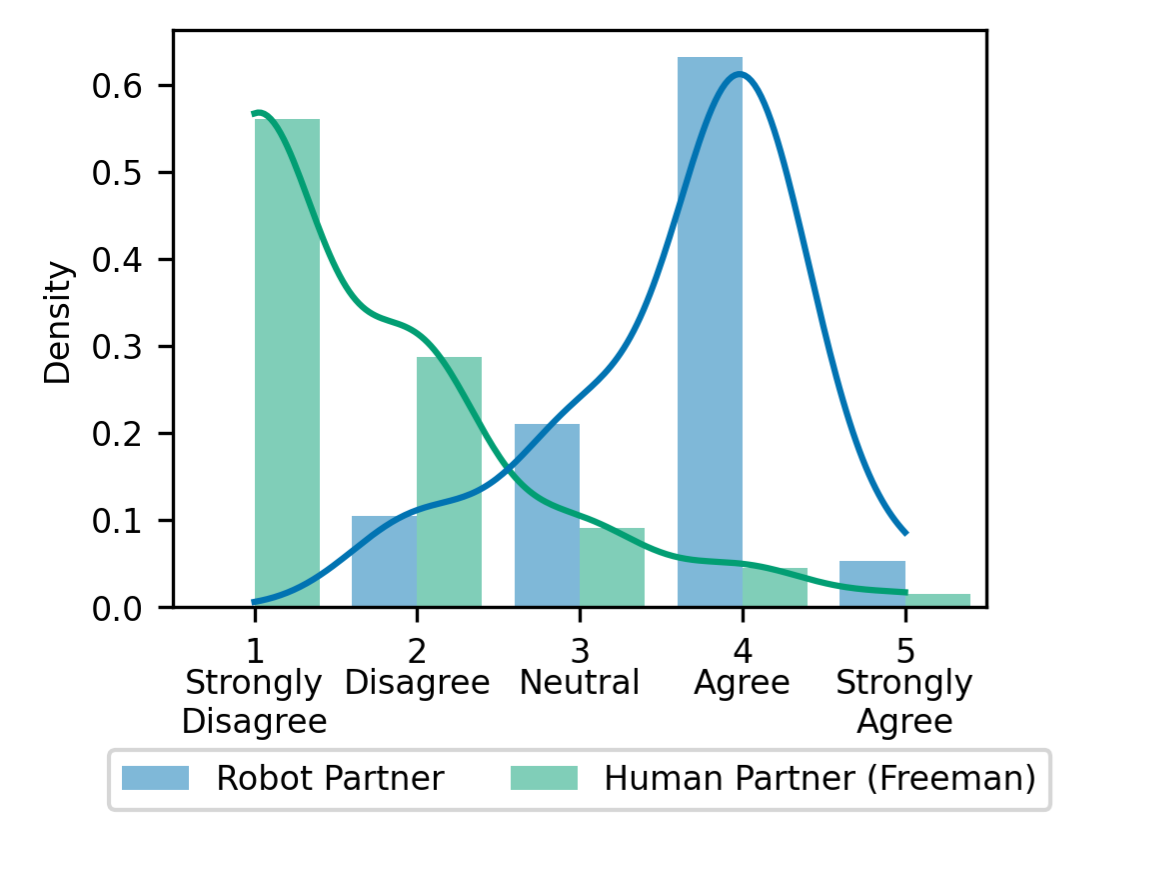}\label{fig:qual-slow}} 
    \hfill 
    \caption{Post-experiment survey responses comparing human partners from Freeman et al. \cite{freeman_classification_2024}, Shaw \cite{shaw_thesis_2024}, and our robot partner on a 5-point Likert scale.}
    \label{fig:qual-likert}
    \Description[Comparison of survey responses across human-human and human-soft-robot collaboration studies.]{A figure containing four bar plots displaying survey responses. Each plot compares responses across different studies on a 5-point Likert scale. Plot (a) compares perceived partner reliability across HH-F, HH-S, and HSR conditions. Plot (b) shows ratings of interaction smoothness across all three conditions. Plot (c) displays perceived freedom of movement across all conditions. Plot (d) compares speed satisfaction between HH-F and HSR conditions only.}
\end{figure}

Consistent with the pre-study findings, the post-study comparisons in Table~\ref{tab:qual-pre-post} further reinforce that human partners ranked significantly higher in nearly every metric. While this statistical difference indicates superior performance with human partners, it should not be interpreted as poor robot performance. In fact, for several key metrics, our robot's performance showed no significant differences from human partners in Shaw's study. When rating partner reliability (Figure~\ref{fig:qual-rely}), participants scored our robot (M = 3.58, SD = 0.90) similarly to Shaw's human partners (M = 3.92, SD = 0.89; p = 0.105), though both were rated lower than Freeman's human partners (M = 4.70, SD = 0.68; p < 0.001) where both participants had complete knowledge of the goal and object state. For object handling confidence, our robot (M = 4.68, SD = 4.68) achieved nearly identical ratings to Shaw's human partners (M = 4.70, SD = 0.62; p = 0.572), approaching the high confidence levels seen with Freeman's partners (M = 4.83, SD = 0.45; p < 0.001). Similarly, adaptation speed showed our robot (M = 3.58, SD = 0.90) performed comparably to Shaw's human partners (M = 3.96, SD = 0.77; p = 0.093), though both differed significantly from Freeman's higher ratings (M = 4.82, SD = 0.46; p < 0.001). These findings suggest that while the robot may not have matched the performance of a human partner with full leadership and task knowledge as presented by Freeman's study, it achieved interaction quality comparable to human partners acting as followers as presented in Shaw's. 

Broader comparison of survey responses revealed consistent patterns across studies. Shaw's participants typically rated their partners between the higher scores seen in Freeman's study and the lower scores observed in our robot study. This pattern emerged across multiple metrics, including experience, qualification, skill, competence, satisfaction, responsiveness, and consistency. The distribution of responses for Shaw's human partners often mirrored those given to our robot partner, albeit with a positive shift in magnitude. This contrasted with Freeman's results, which showed more tightly clustered responses at the higher end of the scale. Figure~\ref{fig:qual-smooth}, which reports interaction smoothness, particularly illustrates this pattern of distribution across the three studies.

Perceptions of partner strength on a 5-point Likert scale showed notable shifts between pre- and post-study surveys for both human and robot partners. In Freeman's human-human teams, participants' initial high confidence in their partner's strength (M = 4.63, SD = 0.57) increased further after the interaction (M = 4.89, SD = 0.31), showing a moderate but significant increase (mean shift = 0.27, SD = 0.54; p < 0.001). More dramatically, participants' initial uncertainty about the robot's strength capabilities (M = 3.84, SD = 1.07) transformed into high confidence post-interaction (M = 4.74, SD = 0.45), demonstrating a substantial and significant increase (mean shift = 0.89, SD = 1.15; p < 0.01). This larger shift in robot partner perception suggests that direct interaction effectively addressed initial concerns about the robot's strength capabilities, bringing final confidence levels close to those reported for human partners.

Despite achieving comparable performance in several metrics, there were areas where the robot's performance notably lagged behind both human studies. Movement quality showed particular deficiencies, with participants rating interaction smoothness (Figure~\ref{fig:qual-smooth}) significantly lower with the robot (M = 2.68, SD = 0.82) compared to both Shaw's (M = 3.80, SD = 0.98; p < 0.001) and Freeman's human partners (M = 4.61, SD = 0.52; p < 0.001). Participants also reported less movement autonomy with the robot (M = 2.53, SD = 0.84; Figure~\ref{fig:qual-move}), indicating they were often unable to move as they desired compared to interactions with Shaw's (M = 3.78, SD = 0.89) or Freeman's human partners (M = 4.41, SD = 0.55). Furthermore, while human-human pairs generally maintained satisfactory speeds (M = 1.67, SD = 0.93 for Freeman's study, indicating faster than desired movement was rarely an issue), over 60\% of participants reported that the robot moved slower than they preferred (M = 3.63, SD = 0.76; p < 0.001; Figure~\ref{fig:qual-slow}).

\section{Discussion}
\label{sec:disc}

As expected for a preliminary study, the human-human teams outperformed the human-soft-robot teams for all quantitative and most qualitative metrics. While this performance gap was anticipated, these results establish an important baseline and demonstrate the viability of soft robotic co-manipulation, marking a significant first step toward more advanced human-soft-robot collaboration.

\subsection{Speed}
Analysis of movement velocities (Figure~\ref{fig:hr-vel-hist}) reveals that participants rarely approached the robot's maximum velocity of 0.5 m/s, suggesting that the longer task completion times (Table~\ref{tab:hh-b-completion-time}) and perceived slower motion (Figure~\ref{fig:qual-slow}) cannot be attributed to the robot's speed limitations alone. Several factors may explain this phenomenon. Under the implemented displacement control scheme, participants may have simply maintained conservative arm positions that felt comfortable but resulted in slower movements. Since velocity was proportional to arm displacement, participants might have prioritized comfortable force resistance over achieving desired speeds. This could be addressed in future work by increasing the proportional gain in the displacement control, allowing higher velocities with similar arm displacements. Alternatively, the slower speeds may reflect participants' lower trust in the robot partner, as indicated by our survey results. This suggests that extended familiarization periods might help participants develop greater comfort with the robot, potentially leading to more dynamic interactions that better utilize the robot's full velocity range. Understanding this balance between control comfort and speed efficiency will be crucial for optimizing human-soft-robot co-manipulation in time-sensitive applications like manufacturing, logistics, or search and rescue operations.

\subsection{Strength}
The dramatic improvement in perceived strength capability is particularly notable given the traditional association of soft robots with lighter, more delicate tasks. The post-study confidence levels approaching those of human partners (4.74 vs 4.89) suggest that direct interaction effectively demonstrated the soft robot's ability to safely handle heavy loads while maintaining its inherent compliance advantages. These findings challenge conventional assumptions about soft robot limitations and indicate promising applications where both strength and adaptability are required.

\subsection{Motion}

Both human-human and human-soft-robot teams demonstrated superior performance in longitudinal (forward/backward) movements compared to lateral (left/right) movements. Since this directional preference appears to be inherent to human movement patterns rather than a limitation of the robot system, future development efforts should prioritize improving overall robot performance rather than attempting to equalize performance across movement directions. However, the robot system introduced additional complexities that affected user experience and task execution.


Participants reported issues with the robot's movement fluidity and control responsiveness. These challenges stemmed from both control system design and mechanical characteristics. On the control side, the rectangular deadband in the displacement control meant diagonal motion (at approximately 45°) required greater input displacement compared to orthogonal directions, often resulting in inconsistent motion where the robot would move either forward or sideways instead of diagonally as intended. A circular deadband could address this limitation. 

Additionally, in an effort to avoid tangled wires, the robot's wheels were programmed to reorient themselves if they were about to pass their predetermined angular thresholds. However, this reorientation strategy often caused short delays in the motion of the center of the mobile base, resulting in start-stop behavior. This behavior was most often seen during diagonal motions. While some participants adapted by breaking diagonal paths into separate forward and sideways movements—a strategy also observed in the human-human study—the mechanical inefficiencies in the wheel mechanism likely made this decomposition more necessary. Video of this behavior is included as supplemental material and can be found at \hyperlink{https://youtu.be/Akah0OdJmCg}{https://youtu.be/Akah0OdJmCg}.


The system also exhibited intrinsic mechanical biases in the arm that slightly affected its performance. Despite identical manufacturing, the bellows chambers showed varying resting lengths due to natural variations in their equilibrium states and loading history. While compressive preloading helped mitigate these differences, the arm retained directional bias in its compliant behavior, responding non-uniformly to applied forces. This asymmetric compliance occasionally manifested as lateral drift during straight-line longitudinal motion.


These combined control and mechanical factors likely contributed to the lower ratings for movement smoothness and control intuitiveness compared to human-human teams. However, the observation that participants could develop compensatory strategies suggests that improved control algorithms and mechanical design could significantly enhance the system's motion performance while maintaining its inherent compliance advantages.


\section{Future Work}
Several promising directions emerge from this study's findings. First, to address the velocity limitations, future work should investigate the displacement controller gains and overall strategy to ensure higher velocities while maintaining comfortable arm displacements. Further consideration should be given to the effect of extended familiarization time periods on user trust and movement speeds.

The robot's motion control system presents multiple opportunities for improvement. Implementing a circular deadband would create more intuitive diagonal movements by equalizing the required displacement across all directions. The current reactive control system could be enhanced with predictive or model-based controllers to reduce the start-stop behavior during direction changes, particularly for diagonal movements. Additionally, more sophisticated compliance compensation algorithms could help mitigate the intrinsic mechanical biases observed in the bellows system.

On the mechanical side, future iterations should focus on improving the bellows design to achieve more consistent resting lengths and uniform compliance behavior. This could involve exploring alternative manufacturing processes or materials that provide more predictable mechanical properties while maintaining the advantages of soft robotics.

Given the encouraging results regarding perceived strength capabilities, future studies should explore expanding the range of collaborative tasks, particularly in applications requiring both robust force handling and inherent safety. This could help further challenge traditional assumptions about soft robot limitations.

\section{Conclusion}
\label{sec:conclusion}

This work presented a novel attempt at large-scale human-soft-robot co-manipulation. A robotic platform consisting of an omni-directional base and a compliant, three-link bellows manipulator was introduced. A benchmark study was performed wherein a human and robot collaborate to move a stretcher like object through simple translational tasks. The data from the study, both quantitative and qualitative, was compared to previous human-human co-manipulation data. While human partners continue to show better results, the gap in performance is relatively small for an initial attempt and can potentially be attributed to small inefficiencies in the motion of the mobile base, or the parameters of the controller. Participants found the robot to be safe, reliable, and capable of supporting its end of the co-manipulated object. Participants' confidence in the robot's strength is particularly notable for a soft robot and demonstrates the promising potential of soft robots' applied to the task of human-robot co-manipulation.


\begin{acks}
This work was funded by the United States National Science Foundation's National Robotics Initiative under grant number 2024792.
\end{acks}

\bibliographystyle{ACM-Reference-Format}
\bibliography{library}










\end{document}